\pdfoutput=1
\documentclass[11pt]{article}

\usepackage{acl}
\usepackage{amsmath}
\usepackage{times}
\usepackage{latexsym}
\usepackage{xcolor}
\usepackage{enumitem}
\usepackage{amssymb}
\usepackage{array}
\usepackage{graphicx}
\usepackage{subcaption}
\usepackage{pifont}
\usepackage{multirow}
\usepackage{bbding}
\usepackage{tcolorbox}
\usepackage{makecell}
\usepackage{tabularx}
\usepackage{float}
\usepackage{fontawesome}
\usepackage{hyperref}
\usepackage[T1]{fontenc}
\usepackage[utf8]{inputenc}

\usepackage{microtype}
\usepackage{booktabs}

\usepackage{inconsolata}
\NewDocumentCommand{\yue}
{ mO{} }{\textcolor{blue}{\textsuperscript{\textit{Yue Huang}}\textsf{\textbf{\small[#1]}}}}
\NewDocumentCommand{\cdp}
{ mO{} }{\textcolor{blue}{\textsuperscript{\textit{Dong ping}}\textsf{\textbf{\small[#1]}}}}
\NewDocumentCommand{\gcj}
{ mO{} }{\textcolor{blue}{\textsuperscript{\textit{Chu jie}}\textsf{\textbf{\small[#1]}}}}

\title{LLM-as-a-Coauthor: Can Mixed Human-Written and Machine-Generated Text Be Detected?}
\author{
\textbf{Qihui Zhang}$^{1}$\footnotemark[1]\footnotemark[2], \textbf{Chujie Gao}$^{1}$\footnotemark[1]\footnotemark[2], \textbf{Dongping Chen}$^{2}$\footnotemark[1], \textbf{Yue Huang}$^{3}$, \textbf{Yixin Huang}$^{4}$,\\
\textbf{Zhenyang Sun$^{1}$\footnotemark[2], Shilin Zhang$^{1}$\footnotemark[2], Weiye Li$^{1}$\footnotemark[2], Zhengyan Fu$^{1}$\footnotemark[2], Yao Wan$^{2}$, Lichao Sun$^{1}$\footnotemark[3] } \\ \\
$^{1}$\textbf{Lehigh University}, $^{2}$\textbf{Huazhong University of Science and Technology},\\
$^{3}$\textbf{University of Notre Dame}, $^{4}$\textbf{Institut Polytechnique de Paris} \\
\texttt{\{maskhui1003, gaochujie1107, dongpingchen0612, james.lichao.sun\}@gmail.com}
}

\begin{document}
\maketitle

\renewcommand{\thefootnote}{\fnsymbol{footnote}}
\footnotetext[1]{Equal contribution.}
\footnotetext[2]{Visiting Students at LAIR Lab, Lehigh University.}
\footnotetext[3]{Lichao Sun is the corresponding author.}

\renewcommand{\thefootnote}{\arabic{footnote}}

\begin{abstract}

With the rapid development and widespread application of Large Language Models (LLMs), the use of Machine-Generated Text (MGT) has become increasingly common, bringing with it potential risks, especially in terms of quality and integrity in fields like news, education, and science. Current research mainly focuses on purely MGT detection without adequately addressing mixed scenarios, including AI-revised Human-Written Text (HWT) or human-revised MGT. To tackle this challenge, we define \textit{mixtext}, a form of mixed text involving both AI and human-generated content. Then, we introduce \textsc{MixSet}, the first dataset dedicated to studying these mixtext scenarios. Leveraging \textsc{MixSet}, we executed comprehensive experiments to assess the efficacy of prevalent MGT detectors in handling \textit{mixtext} situations, evaluating their performance in terms of effectiveness, robustness, and generalization. Our findings reveal that existing detectors struggle to identify \textit{mixtext}, particularly in dealing with subtle modifications and style adaptability. This research underscores the urgent need for more fine-grain detectors tailored for \textit{mixtext}, offering valuable insights for future research. Code and Models are available at \url{https://github.com/Dongping-Chen/MixSet}.
\end{abstract}

\section{Introduction}

The remarkable advancement of Large Language Models (LLM) has sparked global discussions on the effective utilization of AI assistants \citep{openai2022gpt35, openai2023gpt4}. Given that LLMs can correctly follow human instructions and produce useful texts efficiently, more and more people prefer to integrate these powerful tools into their workflow by revising Machine Generated Text (MGT) or using LLMs to polish their Human Written Text (HWT), such as fact-checking revising in journalism \citep{guetta2023chatgpt} and enhancing storytelling in the game industry \footnote{https://aicontentfy.com/en/blog/chatgpt-in-gaming-industry-enhancing-storytelling-and-interaction}.

\begin{figure}[t]
    \centering
    \includegraphics[width=1\linewidth]{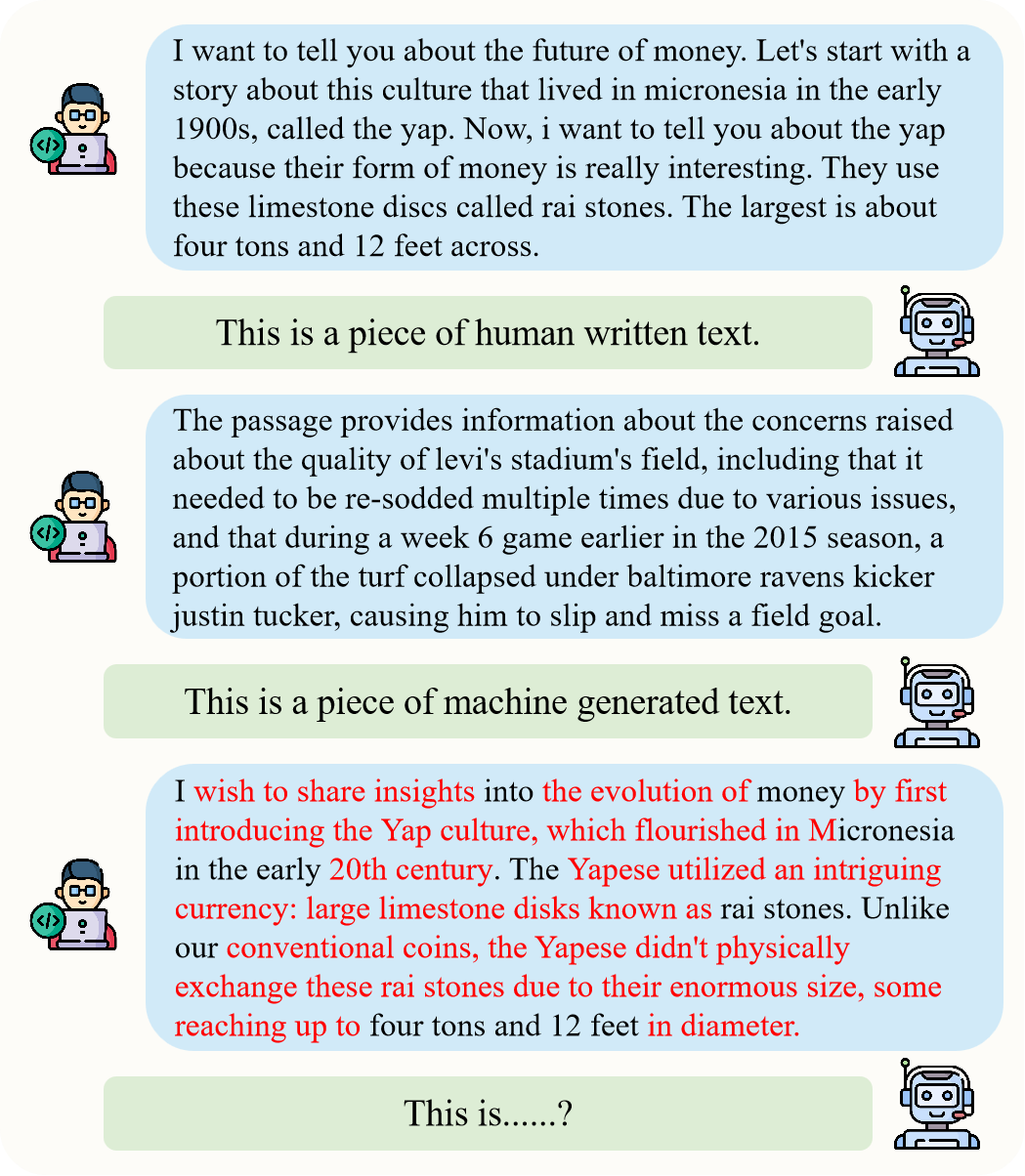}
    \caption{Three kinds of text: Machine Generative Text (MGT), Human Written Text (HWT), and \text{mixtext}. The text come from users \includegraphics[width=0.4cm]{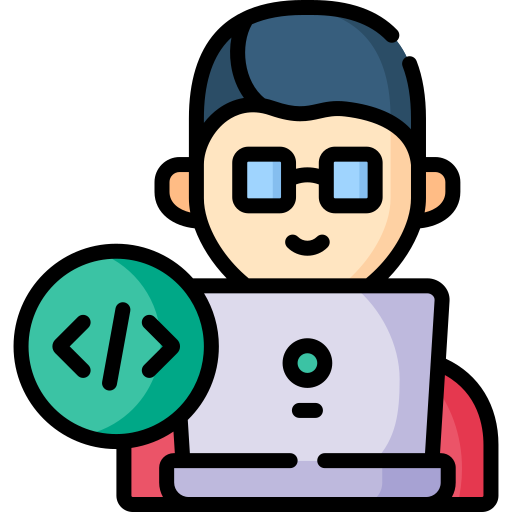} is classified by detectors \includegraphics[width=0.4cm]{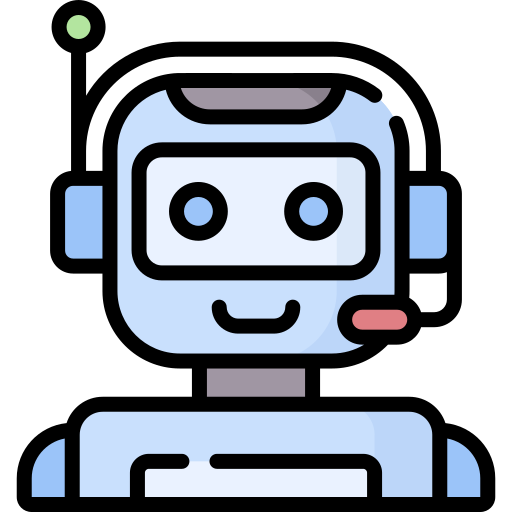}. The text in \textcolor{red}{red} is the HWT polished by LLMs.}
    \label{fig:introduce_mixcase}
\end{figure}

\begin{figure}[h]

    \includegraphics[width=0.48\textwidth]{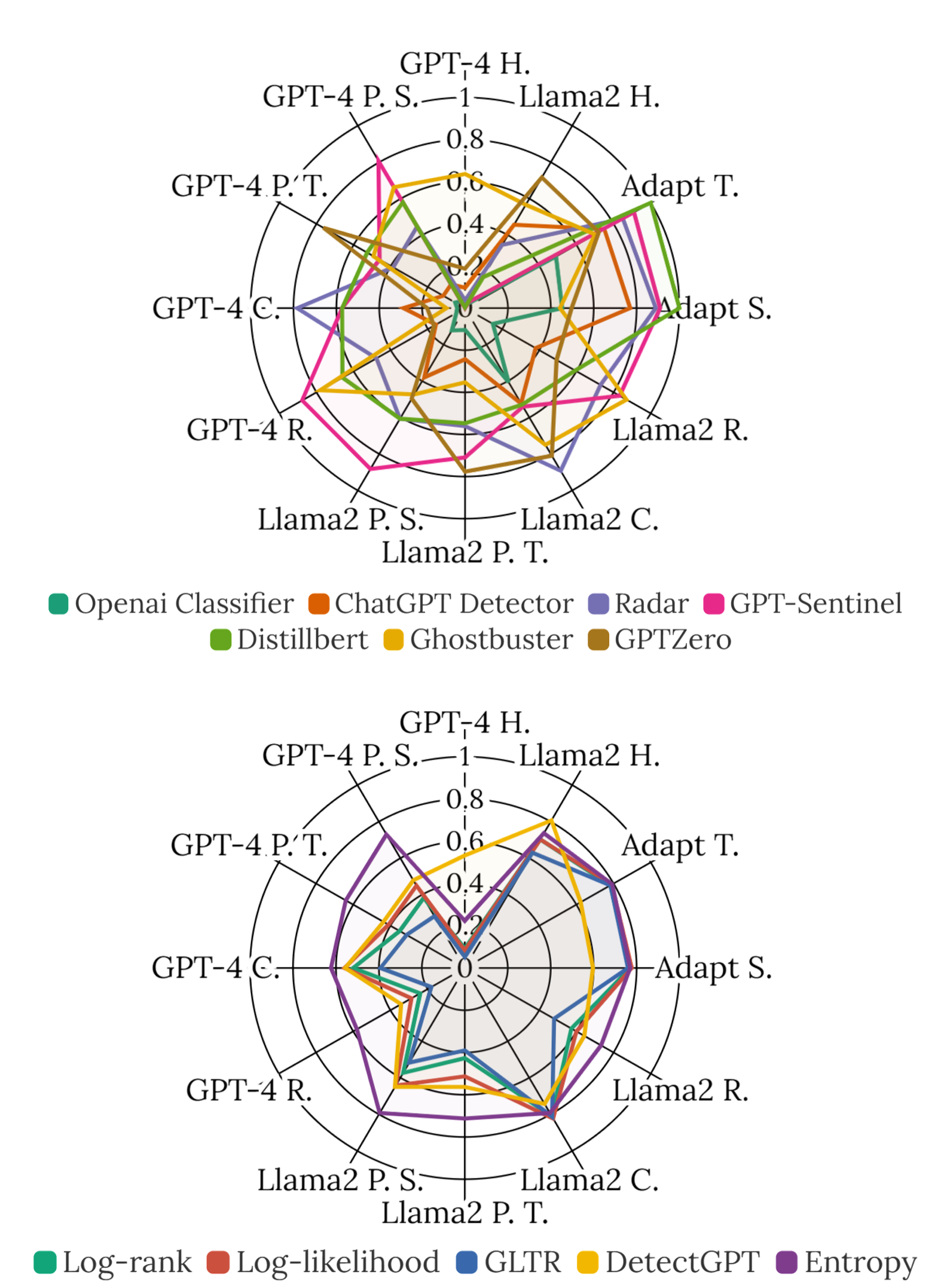}
    \caption{Accuracy of different dectors on \textsc{MixSet}. (Above) Model-based methods; (Below) Metric-based methods. P.T. and P.S. signify token and sentence-level polish, respectively; C. for complete, R. for rewrite; Adapt T. and Adapt S. for token and sentence-level adapt. See \ref{Sec:Section 3} for details on revising operations. }
    \label{fig:ex1_res}
\end{figure}

Despite its various usages, The application of LLMs also causes the potential risk of MGT usage, raising public concerns on various misuse, as seen in the undermining of journalistic integrity and quality \citep{christian2023cnet}, reproducing and amplifying biases \citep{sison2023chatgpt}, plagiarism among students \citep{will2023chatgpt}, and leading disruptions in trust towards scientific knowledge \citep{else2023abstracts}. The misuse of machine-generated text has been a serious concern that is also raised by experts in different fields of work \footnote{https://www.atlantanewsfirst.com/2023/01/24/experts-warn-about-possible-misuse-new-ai-tool-chatgpt/}.

% \yue{This paragraph cannot convey our motivation well: 1) Previous studies (I think) are mainly focusing on two-class detection (purely MGT or HWT). Hence, there is no need to mention "little post-revised texts" and "which only covers one category in mixcase", which may confuse readers. 2) The motivation of our paper is a) The weakness of current research including detection methods and datasets. b) The mixcase is a really important scenario in daily life and has not been well studied. 3) I suggest replacing "post-revised" with "revised", which have the same meaning and are easy to understand. 4) The importance of mixcase text is not well conveyed, can you provide a more clear, understandable example or case? Your examples in this paragraph are mainly about how to use mixcase to evade current detection, which is considered bad behavior, however, I think there are some other benign cases when people need mixcase.}
 
Previous studies proposed many methods to detect MGT, including metric-based and model-based methods, where they have only tried to enhance the detection ability on binary classification, i.e., pure MGT or HWT. However, they did not pay much attention to revised texts (i.e., \textit{mixtext}), but considered these cases as an attack on the detection system \citep{krishna2023paraphrasing} or complex cases for detection \citep{mitchell2023detectgpt, guo2023authentigpt}. However, the mixture of MGT and HWT is an essential scenario in our daily lives when using LLM assistants. For instance, thousands of non-native English speakers utilize LLMs to polish their drafts to avoid grammar problems. Moreover, LLMs can follow human instructions to produce new stories and interactive dialogue in game design \footnote{https://aicontentfy.com/en/blog/chatgpt-in-gaming-industry-enhancing-storytelling-and-interaction}. Authors can also use LLMs to complete stories, providing them with new ideas and inspiration with LLM assistants like \textit{Metaphoria} \cite{Metaphoria} and \textit{Sparks}, thereby generating metaphorical and science writing suggestions and supporting creative writing tasks \cite{MixcaseComplete2}.

% \cdp{Programmers can easily revise code comments automatically generated by LLMs for higher efficiency.}
% \yue{This paragraph can be integrated with the last one. No need to separate them.}

Hence, there is a pressing demand to comprehensively analyze mixture cases and give a formal definition of them. Given that \textit{mixtext} is a very common case in daily life and its amount continuously increases in NLP areas, it holds significant importance, especially in education. To end this, we propose a new dataset \textsc{MixSet}, which is the first dataset that aims at the mixture of HWT and MGT, including both AI-revised HWT and human-revised MGT scenarios as illustrated in Figure \ref{fig:introduce_mixcase}, which addresses gaps in previous research. Further details of the dataset and definitions can be seen in Section \ref{Sec:Section 3}. We also examine our dataset on mainstream detectors in binary and three-class settings to further analyze and raise concerns about these common but hard-to-detect cases.\\ 
To summarize, our work provides three main contributions: 
 \begin{itemize}[nolistsep, leftmargin=*]
     \item We defined \textit{mixtext}, a form of mixed text involving both AI and human-generated content, providing a new perspective for further exploration in related fields.
     \item We proposed a new dataset \textsc{MixSet}, which specifically addresses the mixture of MGT and HWT, encompassing a diverse range of operations within real-world scenarios, addressing gaps in previous research.
     \item Based on \textsc{MixSet}, we conducted extensive experiments involving mainstream detectors and obtained numerous insightful findings, which provide a strong impetus for future research.
 \end{itemize}

\begin{figure*}[h]
    \centering
    \includegraphics[width=\textwidth]{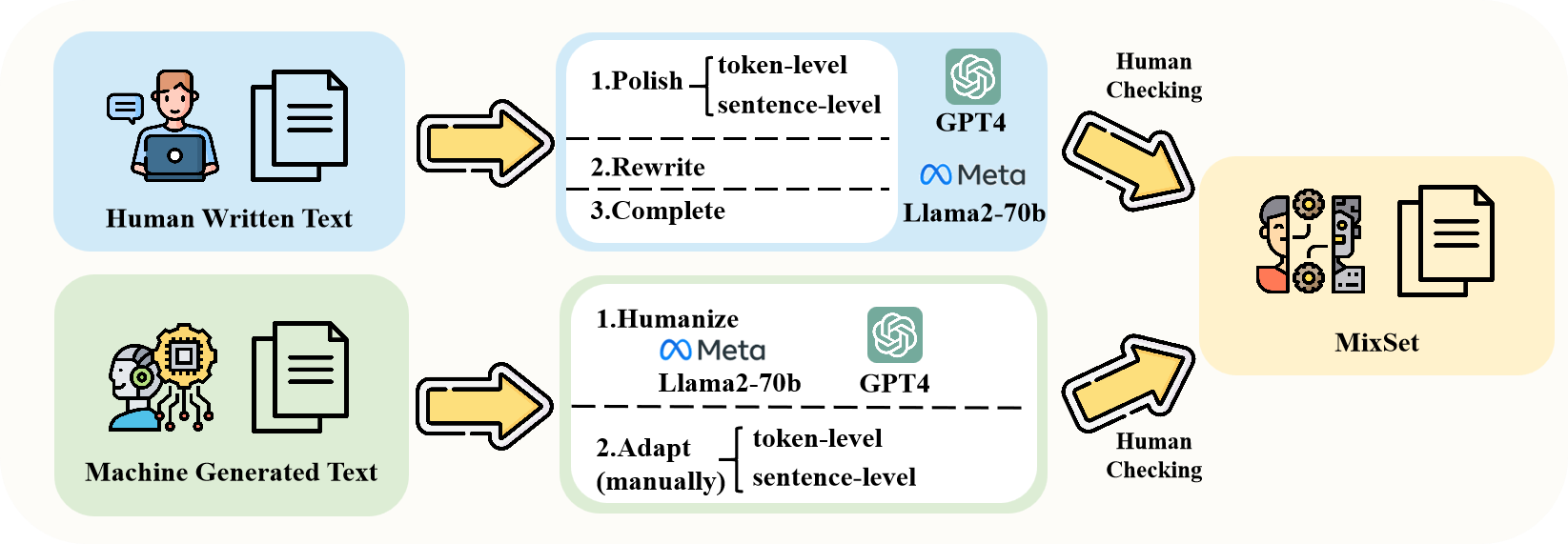}
    \caption{The process of MixSet generation. We perform distinct operations in HWT and MGT. In HWT, three operations—polish, rewrite, and complete—are employed. In MGT, we utilize LLama2 and GPT-4 to aid in humanization and conduct the adaptation operation manually.}
    \label{fig:dataset_generation}
\end{figure*}

\section{Related works}
\subsection{Machine Generated Text Detection}
Current MGT detection methods can be broadly categorized into metric-based and model-based methods according to the previous study \citep{he2023mgtbench}. Please refer to Appendix \ref{sec: appendix A} for comprehensive related works.

\noindent\textbf{Metric-based Methods.} Building upon the observation that MGTs occupy regions with sharp negative log probability curvature, \citet{mitchell2023detectgpt} introduced a zero-shot whitebox detection method called DetectGPT, setting a trend in metric-based detection \citep{su2023detectllm, mireshghallah2023smaller, bao2023fast}. Recently, \citet{yang2023dna} also introduced a powerful detection method known as DNA-GPT, which leverages N-gram \citep{shannon1948mathematical} in a black-box setting. 

\noindent\textbf{Model-based Methods.}
In the era of Large Language Models (LLMs), \citet{guo2023close} developed the ChatGPT Detector based on a fine-tuned Roberta model. As for decoder-based detectors, GPT-sentinel \citep{chen2023gpt} leverage the t5-small model \citep{muennighoff2022crosslingual} and show convincing results when detecting MGT even in revised cases. 

\subsection{Previous study to mix of HWT and MGT}
Prior studies have viewed the mixture of HWT and MGT in different settings. DNA-GPT \citep{yang2023dna} and DetectGPT \citep{mitchell2023detectgpt} notably utilized the T5 model \citep{raffel2020exploring} to simulate scenarios where humans make limited, random modifications to MGT, creating complex test cases. Conversely, DIPPER \citep{krishna2023paraphrasing} and OUTFOX \citep{koike2023outfox} opted for a paraphrasing technique, using this method to craft adversarial attacks aimed at eluding the detection mechanisms of classifiers, thereby presenting a nuanced way to alter MGT while maintaining undetectability.
% Leveraging adversarial learning framework, \citet{hu2023radar} propose a novel detection framework called Radar, which employs an adversarial learning approach to simultaneously train a detector based on Vicuna-7B \citep{vicuna2023} and a paraphrase. 

\subsection{Datasets for MGT Detection}
Previous studies have proposed many datasets of MGT, accompanied by their newly proposed detectors \citep{verma2023ghostbuster, chen2023gpt}. \citet{guo2023close} leverages multiple previous Question-Answer (QA) datasets \citep{jin2019pubmedqa, lin2021truthfulqa}, allowing ChatGPT to generate corresponding answers without explicit prompts. This results in creating a comprehensive dataset comprising a large set of pairs of MGT and HWT. Following the QA pattern, many researchers \citep{mitchell2023detectgpt, su2023detectllm, hu2023radar, he2023mgtbench} propose datasets with the MGT from variant mainstream LLMs \citep{openai2022gpt35, openai2023gpt4}.

However, these datasets typically consist of two distinct classes of texts, namely pure MGT or HWT, without accounting for the potential mixture cases. Furthermore, issues arise due to variations in prompts \citep{koike2023you}, sampling methods, and the inherent differences in length, style, and quality among texts \cite{he2023mgtbench}, posing variations challenges on the generalization ability of proposed detectors \citep{xu2023generalization}. In some instances, MGT included in datasets may not be thoroughly checked, with many noisy sentences not filtered well. For example, some sentences like \textit{Let me know if you have any other questions} exist in the dataset, which will impact the effectiveness of the detectors \citep{guo2023close}. 

\section{\textsc{Mixset} Dataset}
\label{Sec:Section 3}
In this section, we present MixSet (\textbf{\underline{Mix}}case Data\textbf{\underline{set}}), the first dataset featuring a blend of HWT and MGT. Distinguished from earlier datasets exclusively composed of pure HWT and MGT, \textsc{MixSet} comprises a total of $3.6k$ mixtext instances, and the pipeline of its construction is shown in Figure \ref{fig:dataset_generation}. These operations are grounded in real-world application scenarios, each altered by a single LLM or through manual intervention, contributing 300 instances in our \textsc{MixSet}.

For our base data, we meticulously select pure HWT and MGT datasets. In the case of HWT, we gather datasets proposed before the widespread use of LLMs to mitigate potential contamination by MGT, as detailed in Table \ref{HWTdataset}. For MGT, we choose samples from previous datasets \citep{squad, truthfulqa, NarrativeQA_LLMs}, generated in a QA pattern by different LLMs, including the GPT family \citep{openai2022gpt35, openai2023gpt4}, ChatGLM \citep{du2022glm}, BloomZ \citep{muennighoff2022crosslingual}, Dolly \footnote{https://www.databricks.com/blog/2023/04/12/dolly-first-open-commercially-viable-instruction-tuned-llm}, and StableLM \citep{stable2023}, all distinct from our \textsc{MixSet} instances.

\begin{table}[h]
\centering
\small
\setlength{\tabcolsep}{7pt}
\renewcommand\arraystretch{1.3}
\caption{The original resources of Human Written Texts in constructing our \textsc{MixSet}.}
\label{HWTdataset}
\begin{tabularx}{\columnwidth}{cX} 
\toprule[1.5pt]
\textbf{Text Type} & \textbf{Original Resources} \\
\midrule
Email Content & Enron Email Dataset \cite{EnronEmailDataset}\\
News Content & BBC News \cite{BBCNews}\\ 
Game Review & Steam Reviews \cite{steam_reviews_2021}\\
Paper Abstract & ArXiv-10 \cite{arxivPaperAbstract} \\
Speech Content & TED Talk \cite{ted_talk}\\
Blog content & Blog \cite{blogdataset}\\
\bottomrule[1.5pt]
\end{tabularx}
\end{table}

\subsection{Definition of Mixtext}
Generally, \textit{mixtext} is the mixed text involving both AI and human-generated content. To formulate it, we define a text sequence as $x \in X$, where $X$ represents the set of all text sequences. The sequences in $X$ can originate from either human-written text $\mathbb{X}_{\text{human}}$ or machine-generated text $\mathbb{X}_{\text{machine}}$. We denote the set of operations used to revise texts as $\mathbb{OP} = \{OP_1, OP_2, \dots, OP_n\}$, categorized into two groups: ${OP_{\text{human}}, OP_{\text{machine}}}$. Here, $OP_{\text{human}}$ refers to operations involving human revision on machine-generated text (MGT), while $OP_{\text{machine}}$ refers to AI-driven operations on human-written text (HWT). In addition to $\mathbb{X}_{\text{human}}$ and $\mathbb{X}_{\text{machine}}$, we define $\mathbb{X}_{\text{mixtext}}$ as the union of all texts derived from $\mathbb{X}_{\text{human}}$ through $OP_{\text{machine}}$ and all texts derived from $\mathbb{X}_{\text{machine}}$ through $OP_{\text{human}}$:

\begin{align*}
    \mathbb{X}_{\text{mixtext}} = &\{OP_{\text{machine}}(x) \,|\, x \in \mathbb{X}_{\text{human}}\} \\ \cup  &\{OP_{\text{human}}(x) \,|\, x \in \mathbb{X}_{\text{machine}}\}
\end{align*}

\subsection{Dataset Construction}
\label{set: dataset}

Combined with previous studies \cite{textOperation, wang2021adversarial} and real scenarios, we use five operations to generate mixtexts. They are divided into two operations shown in Table \ref{tab: operation}: 1) AI-revised: it contains three operations including `polish', `complete', and `rewrite'. 2) Human-revised: it includes `adapt' and `humanize'.

\begin{table}[]
\centering
\setlength{\tabcolsep}{5pt}
\renewcommand\arraystretch{1.1}
\caption{Different operations with their operation levels. \ding{52} demonstrate that \textsc{MixSet} contains a subset operates at that level.}
\label{tab: operation}
\scalebox{0.85}{
\begin{tabular}{cccc}
\toprule[1.5pt]
% \multicolumn{2}{c}{Operation}    
\textbf{Operation} & \textbf{Token} & \textbf{Sentence} & \textbf{Paragraph} \\
\midrule
% \multirow{3}{*}{\rotatebox[origin=c]{90}{\scriptsize{AI-Revised}}}
 AI-Polish  & \ding{52}     & \ding{52}    &   \ding{56}  \\
 AI-Complete  & \ding{56}  & \ding{56}    & \ding{52}  \\
 AI-Rewrite   & \ding{56}  &   \ding{56}   & \ding{52}  \\
\midrule
% \multicolumn{1}{c}{\multirow{2}{*}{\rotatebox[origin=c]{90}{\scriptsize{Human-Revised}}}}
 Human-Adapt    & \ding{52}   & \ding{52}   & \ding{56}  \\
% \multicolumn{1}{c}{}  
 Humanize & \ding{52}     & \ding{52}    & \ding{52}\\
\bottomrule[1.5pt]
\end{tabular}}

\end{table}

\begin{itemize}[nolistsep, leftmargin=*]
    \item \textbf{Polish} \citep{polish}: Polish operation contains token-level and sentence-level polishing. Token-level makes alterations at the individual word level, including changes such as adjusting words for precision or correcting spelling errors. On the other hand, sentence-level aims to enhance the overall coherence and clarity of the text by revising and restructuring the complete sentence.
    \item \textbf{Complete} \citep{complete}: Complete operation involves taking 1/3 of every text and employing LLMs to generate the rest of the text.
    \item \textbf{Rewrite} \citep{shu2023rewritelm}: Rewrite operation requires LLMs to initially comprehend and extract key information from the given HWT and then rewrite them.
    \item \textbf{Humanize} \citep{humanize}: Humanize operation typically refers to the modification of MGT to more closely mimic the natural noise for LLM \cite{wang2021adversarial} that human writing always brings. We employed LLMs to introduce various perturbations to the pure MGT, including \textit{typo, grammatical mistakes, links}, and \textit{tags}.
    \item \textbf{Adapt} \cite{MixcaseComplete2}: Adapt operation refers to modifying MGT to ensure its alignment to fluency and naturalness to human linguistic habits without introducing any error expression. The adapt operation is also divided into token-level and sentence-level adaptation. We accordingly performed manual annotations on the pure MGT dataset at both the token and sentence levels.
\end{itemize}

The detailed distribution of each category in \textsc{MixSet} is shown in Table \ref{tab: detailed distribution}. All data generated from GPT-4 (300 items) and Llama2 (300 items) have undergone rigorous manual review and modification in the `humanize' operation.
For AI-revised \textit{mixtext} generation, Llama2-70b and GPT-4 were used, both set to default parameters, including a temperature of 1. These models are chosen for their ability to produce high-quality, grammatically correct texts \cite{open_llm_leaderboard}. In human-revised operation, we leverage two LLMs to assist with `humanize' operation. For the adapt operation, we invite eight human experts with excellent English skills to revise MGT carefully to align it with human expression better. The details of human annotation guidelines and prompt template are shown in Appendix \ref{sec: appendix B.1} and \ref{sec: appendix E}. After collecting all revised texts, we conducted a manual evaluation involving data filtering and cleaning to ensure \textsc{MixSet} is high quality, such as removing conversational phrases like `Sure! Here's a possible completion'.

\begin{table}[!ht]
% \vspace{-5pt}
\centering

\setlength{\tabcolsep}{3.6pt}
\renewcommand\arraystretch{1.1}
\caption{Detailed distribution of different operations in \textsc{MixSet}. }
\label{tab: detailed distribution}
\scalebox{0.85}{
\begin{tabular}{cccccc}
\toprule[1.5pt]  
\multicolumn{3}{c}{\textbf{Operation}} & \textbf{GPT-4} & \textbf{Llama2} & \textbf{Human} \\
\midrule

\multirow{4}{*}{\rotatebox[origin=c]{90}{\textbf{AI}}}&  \multirow{4}{*}{\rotatebox[origin=c]{90}{\textbf{Revised}}} & Polish Tok.  & 300 & 300 & —  \\

& &Polish Sen.  & 300 & 300 & —   \\
& &Complete  & 300 & 300 & —  \\
& &Rewrite   & 300 & 300 & —  \\
\midrule

\multirow{3}{*}{\rotatebox[origin=c]{90}{\textbf{Human}}} &  \multirow{3}{*}{\rotatebox[origin=c]{90}{\textbf{Revised}}} & Humanize &  —   & — & (300+300) \\

& &Adapt Tok.  &  —  & —  & 300  \\
& &Adapt Sen.  &  —  & —  & 300  \\
\bottomrule[1.5pt]
\end{tabular}}

\end{table}

\subsection{Dataset Analysis}
Our comprehensive analysis of the \textsc{MixSet} dataset covers length distribution, self-BLEU \citep{zhu2023texygen}, Levenshtein distance \citep{66levenshtein}, and cosine similarity. We only show analysis of length distribution and cosine similarity analysis here; for self-BLEU and POS distribution, refer to Appendix \ref{sec: appendix B2}.

\begin{figure}[h]
        \centering
        \includegraphics[width=0.45\textwidth]{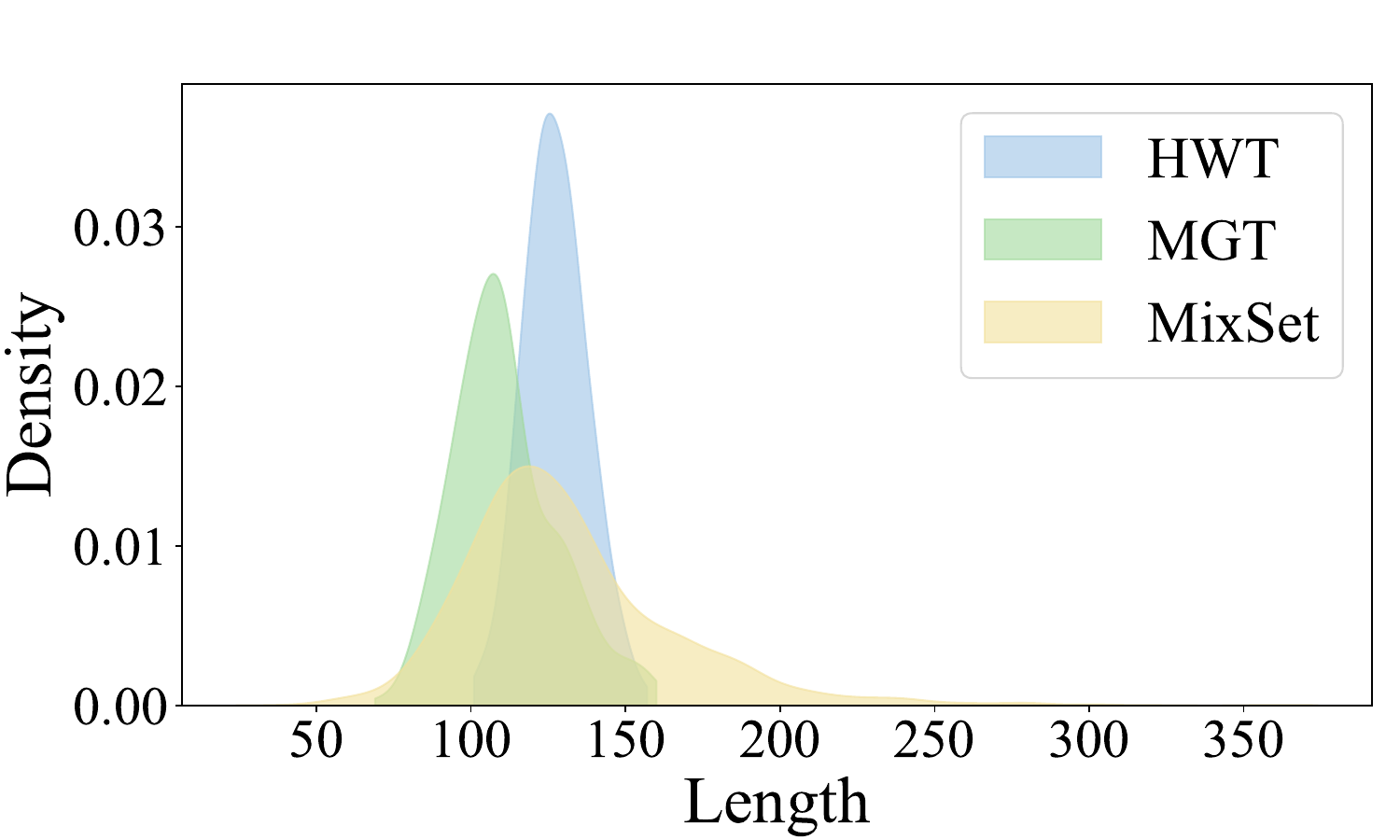}
        \caption{Length distribution of the HWT, MGT, and MixSet.}
        \label{fig:length}
\end{figure}

\begin{figure*}
    \centering
    \includegraphics[width=\linewidth]{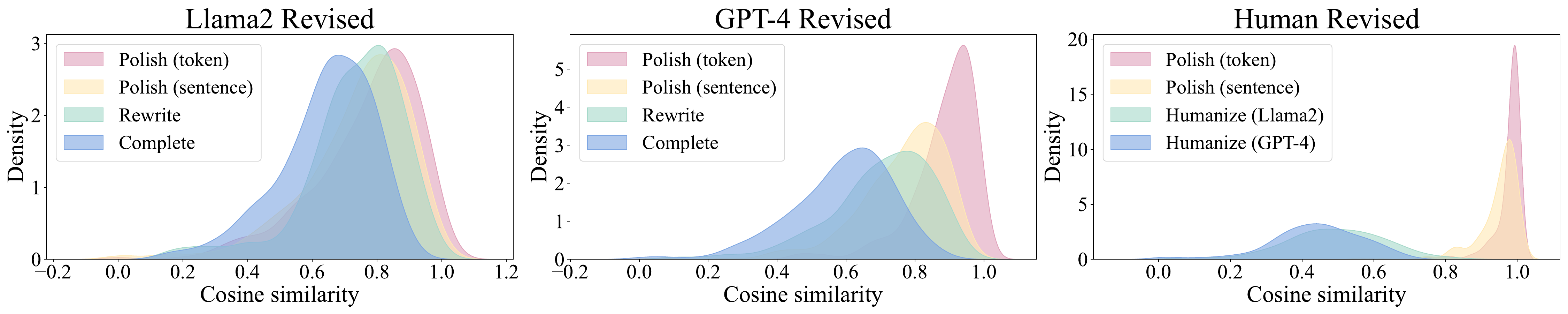}
    \caption{Cosine similarity of the MixSet}
    \label{fig:cosine_similarity}
\end{figure*}

\begin{figure*}[h]
    \centering
    \includegraphics[width=\linewidth]{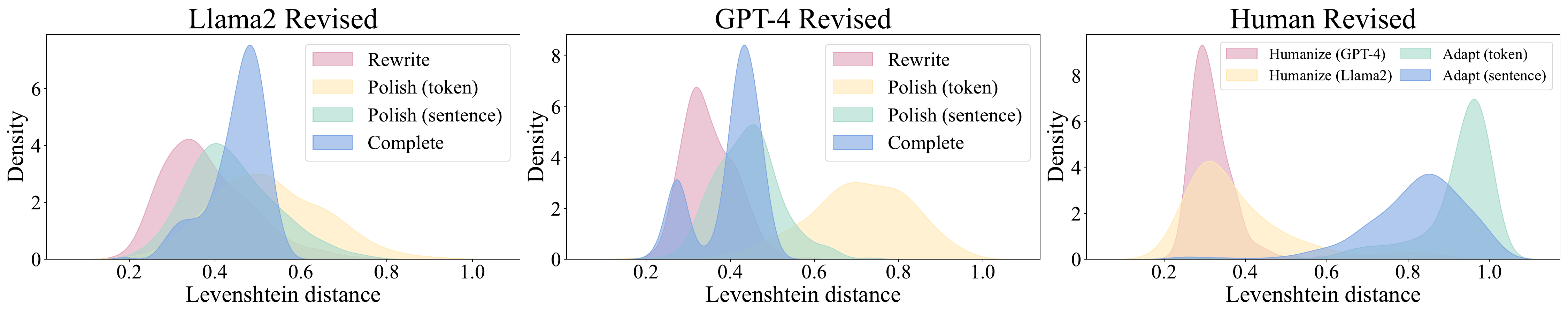}
    \caption{Levenshtein distance of the MixSet}
    \label{fig:Levenshtein}
\end{figure*}

\begin{itemize}[nolistsep, leftmargin=*]
     \item \textbf{Length distribution:} Given that detectors generally perform better with medium to long texts than with short texts \citep{he2023mgtbench}, and to ensure that the text lengths in the \textsc{MixSet} reflect real-world usage patterns, we have systematically selected data with a word count that falls within the range of 50 to 250 words. This range was chosen to ensure that the data were sufficiently detailed to provide meaningful insights while being concise enough to allow for effective analysis and comparison. As shown in Figure \ref{fig:length}, the text lengths of both the \textsc{MixSet}, as well as the HWT and MGT, follow a normal distribution.
     \item \textbf{Cosine Similarity:} Figure \ref{fig:cosine_similarity} illustrates that the texts processed with token-level polish operations exhibit the highest similarity to the original texts, followed by sentence-level polish, rewrite, and complete. Texts modified through the `humanize' operation demonstrate lower similarity than those altered by adaptation.
     \item \textbf{Levenshtein Distance:} The Levenshtein distance \citep{66levenshtein} is a metric for measuring the difference between two strings. We can observe in Figure \ref{fig:Levenshtein} that in terms of the extent of modification, the rewrite operation results in the most significant alterations to the original texts, followed by complete and sentence-level polish. We also observe that manual annotations at both the token-level and sentence-level adaptation exhibit a high degree of differentiation.
\end{itemize}

\section{Experiments}

\subsection{Goals}
We conduct experiments to understand better multiple facets of current detectors encountering our dataset \textsc{MixSet}, including zero-shot and fine-tuning settings. We will figure out four questions: 
\begin{itemize}[nolistsep, leftmargin=*]
    \item \textbf{Question 1.} How do current detectors perform in \textsc{MixSet} dataset? Is there any classification preference in these detectors? 
    \item \textbf{Question 2.} What is the performance of detectors retrained on our \textsc{MixSet}? What about three-classed classification as we consider \textit{mixtext} as a new class distinct from HWT and MGT?
    \item \textbf{Question 3.} What is the generalization ability of current detectors on our \textsc{MixSet}?
    \item \textbf{Question 4.} Will the size of the training set impact the detection ability on \textit{mixtext}?
\end{itemize}

\subsection{Experiment Setup}

Among our four experiments, We evaluate five metric-based and seven model-based detectors on three metrics in total, as shown in Tabel \ref{tab: detectors} and Table \ref{tab:label_setting}. We also outline the detailed training set construction in Table \ref{tab: train set size}. Please refer to Appendix \ref{sec: appendix B2} for a comprehensive introduction to detectors and metrics.

\noindent\textbf{Class Number.} In real-world scenarios, people often aim to detect the presence of MGT in the text (e.g., spreading fake news or propaganda \citep{christian2023cnet}, reinforcing and intensifying prejudices \citep{sison2023chatgpt}), and sometimes mixtext is also treated as MGT (e.g., student modified some words in MGT (i.e., mixtext) to generate homework, to avoid detection). Therefore, our experiments established two categorization systems: binary and three-class. In the binary classification, mixtext is categorized as MGT, while in the three-class classification, mixtext is treated as a separate class. The label setting is shown in Table \ref{tab:label_setting}. 

% tab: detectors
\begin{table}[ht]

\centering
\small
\setlength{\tabcolsep}{6pt}
\renewcommand\arraystretch{1}
\caption{Detectors used in different experiments.}
\begin{tabular}{cccccc}
\toprule[1.5pt]

\multicolumn{2}{c}{\textbf{Detector}}           & \textbf{Q 1} & \textbf{Q 2} & \textbf{Q 3} & \textbf{Q 4} \\

\midrule

\multirow{10}{*}{\rotatebox[origin=c]{90}{\textbf{Metric-Based}}} & Log-likelihood       & \multirow{2}{*}{{\ding{52}}}       & \multirow{2}{*}{{\ding{52}}}  & \multirow{2}{*}{{\ding{56}}} & \multirow{2}{*}{{\ding{52}}}   \\

& \cite{solaiman2019release} & &\\
 
 & Entropy      & \multirow{2}{*}{{\ding{52}}}    & \multirow{2}{*}{{\ding{52}}}  & \multirow{2}{*}{{\ding{56}}} & \multirow{2}{*}{{\ding{56}}} \\
 
& \cite{gehrmann2019gltr}  & & \\

 & GLTR               & \multirow{2}{*}{{\ding{52}}}  & \multirow{2}{*}{{\ding{52}}} & \multirow{2}{*}{{\ding{56}}} & \multirow{2}{*}{{\ding{52}}}  \\

 & \cite{gehrmann2019gltr}  &  & \\
 
 & Log-rank           & \multirow{2}{*}{{\ding{52}}}  & \multirow{2}{*}{{\ding{52}}} &\multirow{2}{*}{{\ding{56}}} & \multirow{2}{*}{{\ding{56}}} \\

 & \cite{mitchell2023detectgpt} &  & \\

& DetectGPT            & \multirow{2}{*}{{\ding{52}}}    & \multirow{2}{*}{{\ding{52}}} & \multirow{2}{*}{{\ding{52}}} & \multirow{2}{*}{{\ding{52}}}  \\
& \cite{mitchell2023detectgpt} & & \\
    
    \hline
    
\multirow{14}{*}{\rotatebox[origin=c]{90}{\textbf{Model-Based}}}   

&  Radar     & \multirow{2}{*}{{\ding{52}}} & \multirow{2}{*}{{\ding{52}}} &\multirow{2}{*}{{\ding{56}}} & \multirow{2}{*}{{\ding{52}}}\\

& \citep{hu2023radar}   & & \\

& ChatGPT Detector    & \multirow{2}{*}{{\ding{52}}}                  & \multirow{2}{*}{{\ding{52}}}      & \multirow{2}{*}{{\ding{52}}} & \multirow{2}{*}{{\ding{52}}} \\
& \citep{guo2023close}   & & \\

& DistillBert  & \multirow{2}{*}{{\ding{52}}}     & \multirow{2}{*}{{\ding{52}}}  &\multirow{2}{*}{{\ding{52}}} & \multirow{2}{*}{{\ding{56}}}  \\
& \citep{ippolito2019automatic} & & \\

& GPT-sentinel       & \multirow{2}{*}{{\ding{52}}}        & \multirow{2}{*}{{\ding{52}}}   &\multirow{2}{*}{{\ding{56}}} & \multirow{2}{*}{{\ding{52}}} \\
& \citep{chen2023gpt} & & \\

&   OpenAI Classifier  & \multirow{2}{*}{{\ding{52}}}                    & \multirow{2}{*}{{\ding{56}}}   &\multirow{2}{*}{{\ding{56}}} & \multirow{2}{*}{{\ding{56}}}   \\

&\cite{Openai2023classifier} & & \\

& Ghostbuster   & \multirow{2}{*}{{\ding{52}}}      & \multirow{2}{*}{{\ding{56}}}  &\multirow{2}{*}{{\ding{56}}} & \multirow{2}{*}{{\ding{56}}} \\
& \cite{verma2023ghostbuster} & &  \\

& GPTzero               & \multirow{2}{*}{{\ding{52}}}                    & \multirow{2}{*}{{\ding{56}}}    &\multirow{2}{*}{{\ding{56}}} & \multirow{2}{*}{{\ding{56}}}   \\

& \cite{tian2023gptzero} & & \\

\bottomrule[1.5pt]
\end{tabular}
\label{tab: detectors}
\end{table}

\begin{table}[h]
\small
\setlength{\tabcolsep}{2pt}
\renewcommand\arraystretch{1.2}
\caption{The details of class number, metrics, and whether the detectors are retrained in our experiments. Except for Question 2(b), we implement binary classifications i.e., HWT and MGT. Per. stands for Percentage.}
\label{tab:label_setting}
\begin{tabular}{c|ccccc}
\toprule[1.5pt]
\multirow{2}{*}{\textbf{Setting}} & \multirow{2}{*}{\textbf{Q 1}} & \multicolumn{2}{c}{\textbf{Q 2}}         & \multirow{2}{*}{\textbf{Q 3}} & \multirow{2}{*}{\textbf{Q 4}} \\
\cmidrule(lr){3-4}
                              &       & \textbf{(a)} & \textbf{(b)} &                                      &                                      \\
                                     \midrule
\textbf{Class Num.}&\multicolumn{1}{c}{2-Class}          & 2-Class                & 3-Class                & \multicolumn{1}{c}{2-Class}          & \multicolumn{1}{c}{2-Class}         \\ \midrule 
\textbf{Metric} & \multicolumn{1}{c}{MGT Per.} & F1, AUC & F1 & \multicolumn{1}{c}{AUC} & \multicolumn{1}{c}{F1, AUC} \\
\midrule
\textbf{Retrained?} & \ding{56} & \ding{52} & \ding{52} & \ding{52} & \ding{52} \\

\bottomrule[1.5pt]
\end{tabular}
\end{table}

\noindent\textbf{Question 1.} 
Based on \textsc{MixSet}, we evaluate current detectors to determine the classification preferences on mixtext, i.e., Does the detector tend to classify mixtext as MGT or HWT? We calculate the percentage of mixtext samples categorized to MGT in the experiment. For the DistilBERT detector and other metric-based detectors utilizing logistic regression models, we employ a training set comprising 10,000 pre-processed samples of both pure HWT and MGT. For other detectors, we use existing checkpoints \footnote{\url{https://huggingface.co/TrustSafeAI/RADAR-Vicuna-7B}} \footnote{\url{https://github.com/haok1402/GPT-Sentinel-public}} or API \footnote{\url{https://gptzero.me/}} and evaluate them in a zero-shot setting.

\noindent\textbf{Question 2(a).}
Following \textbf{Question 1}, our inquiry is whether the detector can accurately classify \textit{mixtext} as MGT after training on \textsc{MixSet}. We fine-tune detectors on pure HWT and MGT data and a train split set of our \textsc{MixSet} labeled as MGT. 

% For detailed experiment settings, please refer to Appendix \ref{sec: appendix C}.

\noindent\textbf{Question 2(b).}
On the other hand, assuming that \textit{mixtext} lies outside the distribution of HWT and MGT, we conduct a three-class classification task, treating mixtext as a new label. In this scenario, we adopt multi-label training for these detectors while keeping all other settings consistent.

\begin{table}[h]

\small
\centering
\setlength{\tabcolsep}{11pt}
\renewcommand\arraystretch{1.2}
\caption{An outline of detailed training set construction for each experiment. `Ope.' denotes `operation transfer' in Experiment 3, while `LLM' refers to `LLM transfer'.}
\begin{tabular}{c|cc}
\toprule[1.5pt]
Experiment & HWT/MGT         & \textsc{MixSet}             \\ \midrule[1.5pt]
Q 1          & $10k$           & $0$                   \\
Q 2(a)         & $10k$           & $3k$                \\
Q 2(b)         & $10k$           & $3k$                \\
Q 3(Ope.)        & $1k$            & $0.5k$                 \\
Q 3(LLM)        & $5k$            & $1.5k$                \\
Q 4          & $1k/4k/7k/10k$ & $0/1.5k/3k$ \\ \bottomrule[1.5pt]
\end{tabular}
\label{tab: train set size}
\end{table}

\begin{table*}[h]

\renewcommand{\arraystretch}{1.5}
\small
\centering
\setlength{\tabcolsep}{6pt}
\renewcommand\arraystretch{1.3}

\caption{F1 score of experiment 2 (a) and (b). Tok. stands for token level and Sen. stands for sentence level. We \underline{underscore} the best-performing detector and \textbf{bold} the score greater than $0.8$, which we consider as a baseline threshold for detection. }

\scalebox{0.9}{
\begin{tabular}{l|ccccccccccccc}
\toprule[1.5pt]
\multicolumn{1}{c|}{\multirow{5}{*}{\textbf{Detection Method}}} & \multicolumn{1}{c}{\multirow{5}{*}{\rotatebox[origin=c]{90}{\textbf{Average}}}} & \multicolumn{8}{c}{\textbf{AI-Revised}}             & \multicolumn{4}{c}{\textbf{Human-Revised}}          \\

\cmidrule(lr){3-10} \cmidrule(lr){11-14}

\multicolumn{1}{c}{}  &\multicolumn{1}{|c}{}                                & \multicolumn{2}{c}{Complete} & \multicolumn{2}{c}{Rewrite}            & \multicolumn{2}{c}{Polish-Tok.} & \multicolumn{2}{c}{Polish-Sen.} & \multicolumn{2}{c}{Humanize}                              & \multicolumn{1}{c}{\multirow{2}{*}{\rotatebox[origin=c]{90}{Adapt-Sen.}}} & \multicolumn{1}{c}{\multirow{2}{*}{\rotatebox[origin=c]{90}{Adapt-Tok.}}} \\

 \cmidrule(lr){3-4}  \cmidrule(lr){5-6} \cmidrule(lr){7-8} \cmidrule(lr){9-10} \cmidrule(lr){11-12} 
\multicolumn{1}{c|}{}   & \multicolumn{1}{c}{}                               & \rotatebox[origin=c]{90}{Llama2}      & \rotatebox[origin=c]{90}{GPT-4}      & \rotatebox[origin=c]{90}{Llama2} & \multicolumn{1}{c}{\rotatebox[origin=c]{90}{GPT-4}} & \rotatebox[origin=c]{90}{Llama2}        & \rotatebox[origin=c]{90}{GPT-4}        & \rotatebox[origin=c]{90}{Llama2}          & \rotatebox[origin=c]{90}{GPT-4}         & \multicolumn{1}{l}{\rotatebox[origin=c]{90}{Llama2}} & \multicolumn{1}{l}{\rotatebox[origin=c]{90}{GPT-4}} & \multicolumn{1}{c}{}                                & \multicolumn{1}{c}{}                             \\
\midrule[1pt]

\multicolumn{13}{l}{Experiment 2 (a): Binary Classification} \\
\midrule[1pt]

        log-rank & 0.615 & 0.695  & 0.686  & 0.637  & 0.479  & 0.617  & 0.606  & 0.647  & 0.595  & 0.617  & 0.454  & 0.676  & 0.667   \\
        log likelihood & 0.624 & 0.695  & 0.695  & 0.637  & 0.492  & 0.657  & 0.627  & 0.657  & 0.657  & 0.637  & 0.386  & 0.676  & 0.667   \\
        GLTR & 0.588 & 0.686  & 0.647  & 0.606  & 0.441  & 0.574  & 0.585  & 0.637  & 0.540  & 0.617  & 0.400  & 0.657  & 0.667   \\
        DetectGPT & 0.635 & 0.715  & 0.651  & 0.656  & 0.560  & 0.632  & 0.587  & 0.657  & 0.632  & 0.692  & 0.587  & 0.641  & 0.609   \\
        Entropy & 0.648 & 0.690  & 0.671  & 0.681  & 0.613  & 0.681  & 0.671  & 0.681  & 0.671  & 0.623  & 0.430  & 0.681  & 0.681   \\ \hline

        Openai Classifier & 0.209 & 0.171 & 0.359 & 0.031 & 0.197 & 0.145 & 0.270 & 0.247 & 0.439 & 0.247 & 0.316 & 0.000 & 0.090  \\

        ChatGPT Detector & 0.660 & 0.705  & 0.696  & 0.676  & 0.583  & 0.676  & 0.647  & 0.647  & 0.594  & 0.667  & 0.615  & 0.705  & 0.705  \\
        
        Radar & \underline{\textbf{0.876}} & \underline{\textbf{0.867}}  & \underline{\textbf{0.877}}  & \underline{\textbf{0.877}}  & \underline{\textbf{0.877}}  & \underline{\textbf{0.877}}  & \underline{\textbf{0.877}}  & \underline{\textbf{0.877}}  & \underline{\textbf{0.877}}  & \underline{\textbf{0.877}}  & \underline{\textbf{0.877}}  & \underline{\textbf{0.877}}  & \underline{\textbf{0.877}}   \\
        GPT-sentinel & 0.713 & 0.714  & 0.714  & 0.714  & 0.714  & 0.714  & 0.714  & 0.714  & 0.714  & 0.696  & 0.714  & 0.714  & 0.714   \\
        Distillbert &0.664 & 0.667  & 0.667  & 0.667  & 0.667  & 0.667  & 0.667  & 0.667  & 0.667  & 0.639  & 0.667  & 0.667  & 0.667  \\
        
\midrule[1pt]
\multicolumn{13}{l}{Experiment 2 (b): Three-class Classification} \\
\midrule[1pt]

DetectGPT & 0.255 & 0.276 & 0.210 & 0.295 & 0.278 & 0.283 & 0.234 & 0.271 & 0.237 & 0.280 & 0.222 & 0.233 & 0.235  \\ 

ChatGPT Detector & 0.304 & 0.288 & 0.346 & 0.283 & 0.288 & 0.395 & 0.341 & 0.265 & 0.328 & 0.267 & 0.317 & 0.253 & 0.273  \\ 

Radar & \underline{0.775} & \underline{\textbf{0.804}} & \underline{\textbf{0.842}} & \underline{0.797} & \underline{\textbf{0.837}} & \underline{\textbf{0.831}} & \underline{\textbf{0.820}} & \underline{\textbf{0.815}} & \underline{\textbf{0.837}} & \underline{\textbf{0.884}} & \underline{\textbf{0.889}} & \underline{0.510} & \underline{0.429}\\

Distillbert & 0.261 & 0.267 & 0.333 & 0.319 & 0.329 & 0.294 & 0.309 & 0.294 & 0.329 & 0.309 & 0.342 & 0.000 & 0.010  \\

\bottomrule[1.5pt]
\end{tabular}}
\label{tab: F1 Ex2}
\renewcommand{\arraystretch}{1}
\end{table*}

\begin{table}[h]
\small
\centering
\renewcommand{\arraystretch}{1.2}
\caption{The detection capabilities on pure HWT and MGT, comparing performances with (w.) and without (w.o.) MixSet labeling MGT during the training process, with the better one \underline{underscored}.}
\label{tab: retrain detectors capability in binary classification}
\scalebox{1}{
\begin{tabular}{l|cc|cc}
\toprule[1.5pt]
\multirow{2}{*}{\textbf{Detector}} & \multicolumn{2}{c|}{\textbf{F1}} & \multicolumn{2}{c}{\textbf{AUC}} \\
& w.o.  & w. & w.o. & w.  \\
\midrule
log-rank & \underline{0.830} & 0.821 & 0.922 & 0.922 \\
log likelihood & \underline{0.845} & 0.834 & 0.931 & 0.931 \\
GLTR & \underline{0.831} & 0.818 & 0.920 & 0.920 \\
DetectGPT & \underline{0.746} & 0.725 & 0.820 & 0.820 \\
Entropy & 0.770 & 0.770 & 0.859 & 0.859 \\ \midrule
ChatGPT Det. & 0.881 & \underline{0.896} & 0.954 & \underline{0.979} \\
Radar & 0.997 & 0.997 & 1.000 & 1.000 \\
GPT-sentinel & \underline{0.988} & 0.982 & \underline{1.000} & 0.999 \\
Distillbert & \underline{0.996} & 0.984 & 1.000 & 1.000 \\
\bottomrule[1.5pt]
\end{tabular}}

\end{table}

\noindent\textbf{Question 3.}
As highlighted in prior research  \citep{xu2023generalization,he2023mgtbench} that transfer ability is crucial for detectors, our objective is to investigate the effectiveness of transferring across different subsets of \textsc{MixSet} and LLMs. We establish two transfer experiments to assess whether the transferability of current detection methods is closely linked to the training dataset, referred to as operation-generalization and LLM-generalization:

\begin{itemize}[nolistsep, leftmargin=*]
    \item \textbf{Operation-generalization:} We initially train our detectors on one \textsc{MixSet} subset operated by one of these operations, along with pure HWT and MGT datasets, and then proceed to transfer it to the subsets processed by other operations.
    \item \textbf{LLM-generalization:} In this experiment, we train detectors on GPT-generated texts and HWT, following which we evaluate the detectors on mixtext generated by GPT family \citep{openai2023gpt4} and Llama2 \citep{touvron2023llama}, respectively, to see whether there is a generalization gap between different LLMs.
\end{itemize}

\noindent\textbf{Question 4.}
Empirically, incorporating more training data has been shown to enhance detection capabilities and robustness for generalization \citep{ying2019overview}. To determine the relation between detectors' performance and the size of the training set, we follow \textbf{Question 2} and use varying sizes of training sets to retrain detectors, as illustrated in Table \ref{tab: train set size}.

\begin{figure*}
    \centering
    \includegraphics[width=\linewidth]{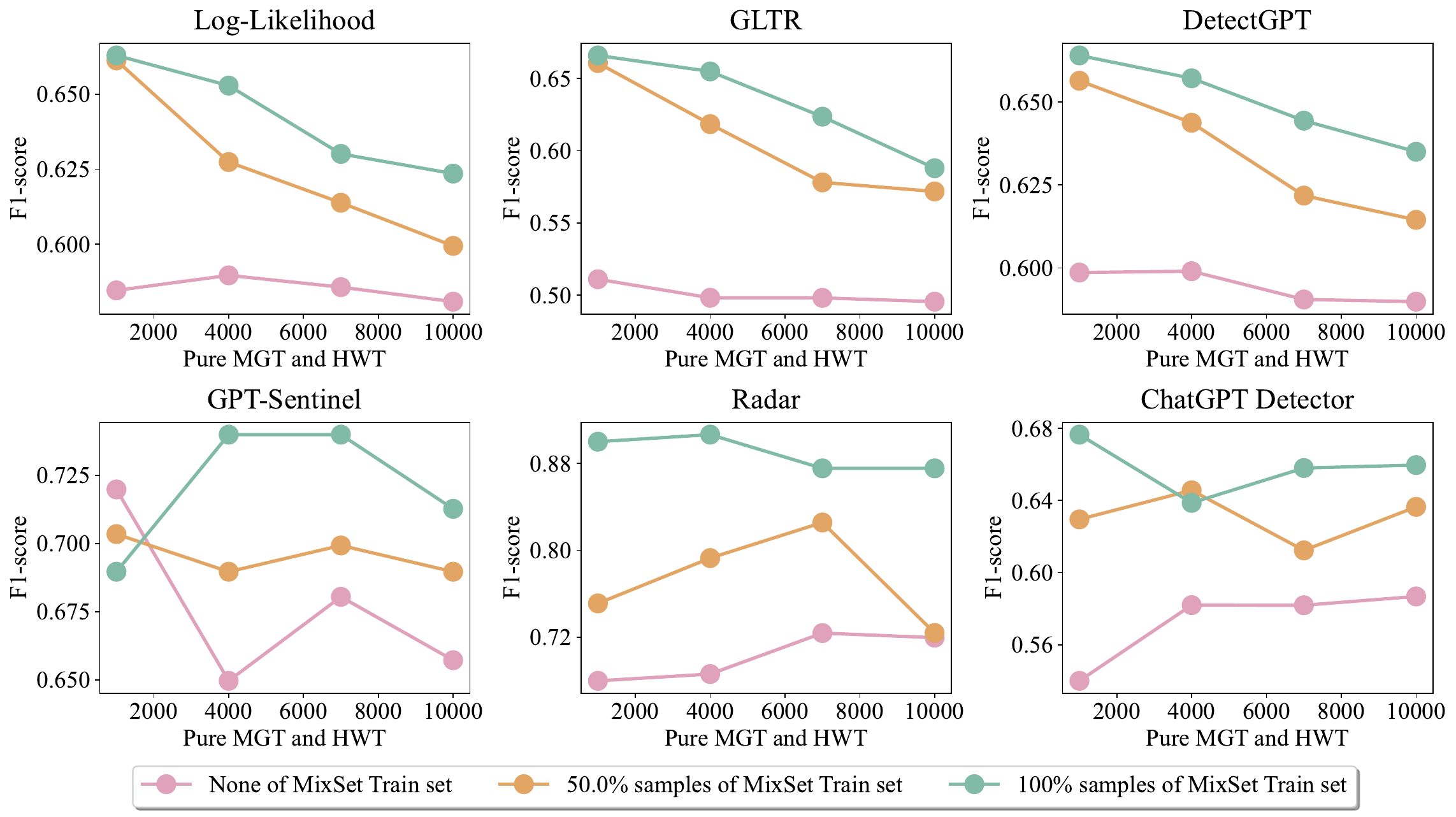}
    \caption{Analysis of the F1-score performance of various detectors across differing quantities of mixtext instances from \textsc{MixSet}, as well as pure MGT and HWT.}
    \label{fig: Ex4}
\end{figure*}

\section{Empirical Findings}
\noindent\textbf{There is no obvious classification preference in current detectors on mixtext.} In other words, the detectors do not exhibit a strong tendency to classify mixtext as either HWT or MGT. As we can observe from Figure \ref{fig:ex1_res} and Table \ref{tab: Ex1 details}, it is evident that the MGT percentage\footnote{MGT percentage means the percentage of identifying samples as MGT of different sets in Experiment 1.} of \textit{mixtexts} is between MGT and HWT, indicating that the current detectors do not have a strong preference towards mixtext classification. This proves the success and effectiveness of our constructed \textsc{MixSet} in presenting mixed features of HWT and MGT, demonstrating the limitations of existing detectors in recognizing mixtext. 

When dealing with mixtext, the detectors treat it as an intermediate state between HWT and MGT. Most detectors exhibit inconsistent classification within a single subset, fluctuating between accuracies of 0.3 and 0.7, akin to random choice. In AI-revised scenarios, subsets, such as polished tokens or sentences, pose extreme detection challenges. Mainstream detectors generally perform poorly in these cases due to the subtle differences between \textit{mixtext} and original text, highlighted in previous studies \citep{krishna2023paraphrasing}. Furthermore, texts generated by Llama2-70b are easier to detect than those by GPT-4, possibly due to GPT-4's closer generative distribution to human writing. 

\begin{figure}[h]
    \includegraphics[width=0.5\textwidth]{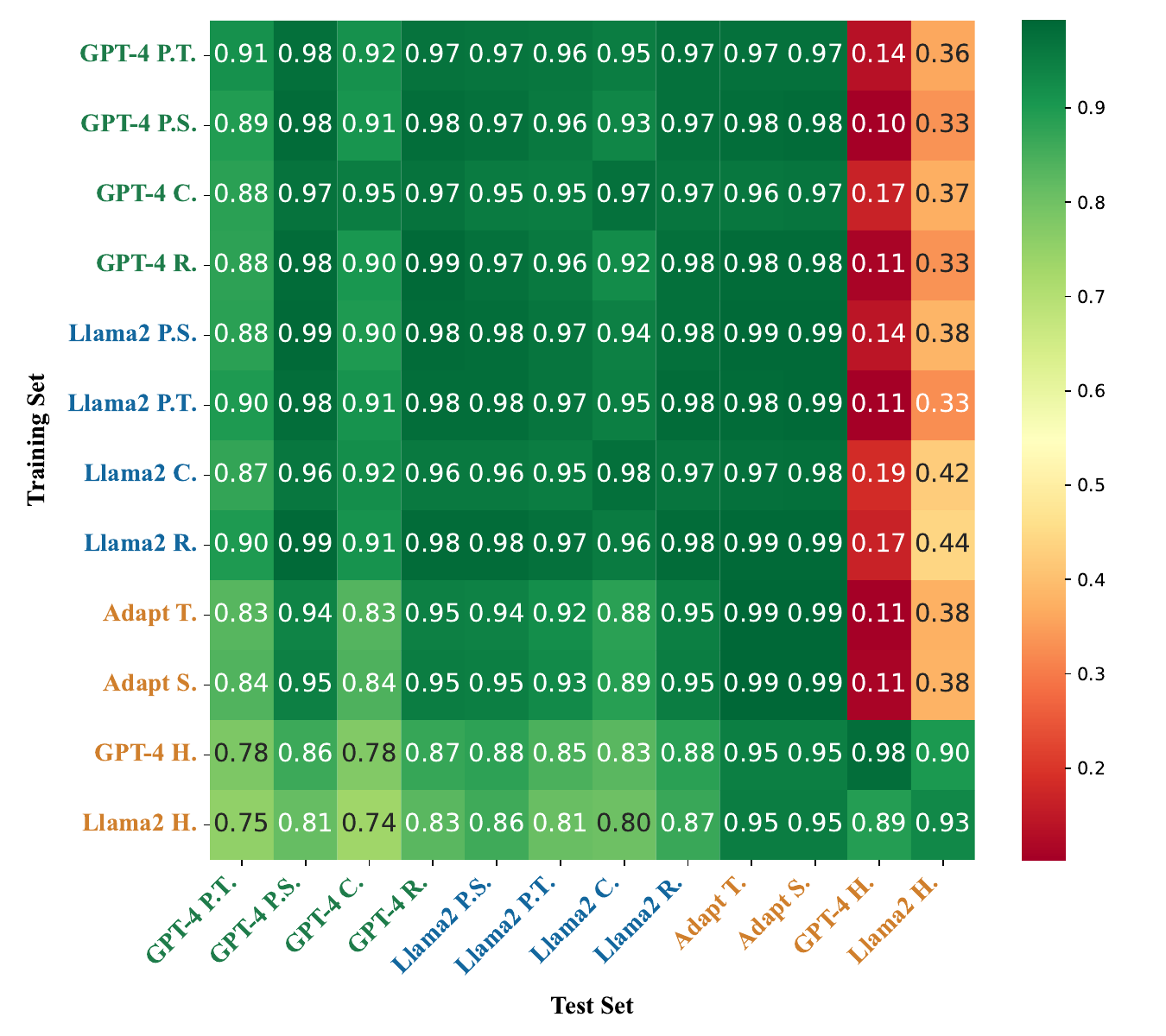}
    \caption{The AUC Heatmap of GPT-sentinel. }
    \label{fig:AUC_gpt_sentinel}
\end{figure}

\noindent\textbf{Supervised binary classification yields profound results; however, three-classes classification encounters significant challenges when applied to mixtext scenarios except Radar.} As indicated in Table \ref{tab: F1 Ex2}, retrained model-based detectors outperform metric-based methods in both binary and three-class classification tasks. Notably, Radar ranks first in our results, achieving a significant lead over other detectors. We suppose that this superior performance can be attributed to its encoder-decoder architecture, which boasts 7 billion trainable parameters, substantially more than its counterparts. We also examined the impact of retraining on MixSet on MGT detection performance. As indicated in Table~\ref{tab: retrain detectors capability in binary classification}, there was a slight decrease in the F1 score, while the AUC metric remained largely unaffected. Notably, post-retraining, the detector acquired the capability to identify mixtext—an advancement deemed highly valuable. This ability to detect mixtext, despite a minor trade-off in F1 score for MGT detection, represents a significant step forward, suggesting a promising direction for enhancing detector versatility and applicability in varied contexts.

In the three-class classification task, detectors based on LLMs, particularly the Radar detector, significantly outperformed those utilizing the BERT model. The BERT-based detectors' performance was markedly poor, akin to random guessing, with some models even underperforming a random baseline. This stark contrast underscores the efficacy of LLMs in capturing nuanced distinctions, as demonstrated in tasks like Mixtext. The superior performance of LLM-based Radar detectors lays a solid foundation for future explorations and applications in fine-grained classification tasks.

\begin{table}[]
\renewcommand{\arraystretch}{1.5}
\small
\centering
\setlength{\tabcolsep}{6pt}
\renewcommand\arraystretch{1.3}

\caption{Result of LLM-transfer experiments. Although we retrain our detector on texts generated by GPT-4, it shows convincing generalization ability to Llama2.
}

\begin{tabular}{l|cc|cc}

\toprule[1.5pt]
\multicolumn{1}{c|}{\multirow{2}{*}{\textbf{Method}}} & \multicolumn{2}{c|}{\textbf{w.o \textit{MixSet}}}& \multicolumn{2}{c}{\textbf{w. \textit{MixSet}}}\\

& Llama2 & GPT-4 & Llama2 & GPT-4 \\

\midrule[1.5pt]

GPT-sentinel & \textbf{0.813} & \underline{0.739} &\textbf{0.972} & \textbf{0.971}  \\

Radar & \underline{\textbf{0.834}} & 0.729 & \underline{\textbf{0.997}} & \underline{\textbf{1.000}} \\
        
ChatGPT Det. & 0.664 & 0.445 & 0.681 & 0.480  \\

Distillbert & 0.687 & 0.638 & 0.673 & 0.698  \\

\bottomrule[1.5pt]
\end{tabular}
\label{tab:LLM transfer}
\renewcommand{\arraystretch}{1}
\end{table}

\noindent\textbf{Current detectors struggle to generalize across different revised operation subsets of \textsc{MixSet} and generative models.}
As shown in Figure \ref{fig:AUC_gpt_sentinel} and Figure \ref{fig:entire_figure}, significant variability is observed in the transfer capabilities of three different detectors. Additionally, training on texts generated by different revised operations results in different transfer abilities for these detectors. Overall, Radar exhibits the most robust transfer capability among the four model-based detectors, achieving an overall classification accuracy exceeding 0.9, followed by GPT-sentinel, DistillBert, and finally, the ChatGPT Detector. Among various operations, `Humanize' exhibits the poorest transfer performance in almost all scenarios. Additionally, other operations also experience significant declines when dealing with `Humanize' mixtexts. This suggests that `Humanize' falls outside the current detectors' distribution of MGT, a gap that could be addressed by retraining on these specific cases. As shown in \ref{tab:LLM transfer} It is also noteworthy that texts generated by Llama2-70b demonstrate stronger transfer abilities than those generated by GPT4.

\noindent \textbf{Increasing the number of \textit{mixtext} samples in the training set effectively enhances the success rate of \textit{mixtext} detection.} However, adding pure text samples does not yield significant improvements and may even have a negative impact on detector performance, especially for metric-based methods. This may be attributed to subtle distribution shifts between mixtext and pure text. The current detector still faces significant challenges in capturing these subtle shifts. For mixtext scenarios, a more powerful and fine-grained detection method is needed.

\section{Conclusion}

In this paper, we defined \textit{mixtext}, the mixed text of human and LLM-generated content. Then, we proposed a dataset \textsc{MixSet} to address the research gap in studying the mixed scenarios of machine-generated text (MGT) and human-written text (HWT). A thorough evaluation of the dataset is conducted, performing binary, three-class, and transfer experiments on mainstream detectors. The results underscore the complexities inherent in identifying mixtext, indicating the challenge of distinguishing the subtle differences in mixtext. As a result, there is a need for more robust and fine-grained detection methods.

\clearpage

\section{Limitation}
\textbf{Bias Introduced by Human Participation.} Although our study involved multiple human participants to modify the text, increasing the diversity and authenticity of the data, the text processing methods of different participants could vary due to their language habits and styles. This might affect the representativeness of the dataset and the generalization ability of the detection models.

\noindent\textbf{Limitation in the Scale of the MixSet Dataset.}  As the MixSet dataset is the first to be proposed for studying mixed texts (mixtext), its overall scale is relatively small despite its comprehensive coverage in types. This could limit the comprehensiveness of model training and evaluation.

\section{Ethics Statement}

\textbf{Opposition to Misuse of Mixed Text Scenarios.} Our study highlights that the mixtext of HWT and MGT could significantly diminish the discerning abilities of detectors. However, we strongly oppose the misuse of mixtext to evade detection in specific scenarios, such as during examinations and homework assignments. We believe such misuse could severely harm the fairness of education and the integrity of academic practices.

\noindent\textbf{Purpose for Scientific Research.} This study aims purely for scientific exploration and understanding of the behavior and impact of mixtext in natural language processing. Our goal is to enhance understanding of mixed text processing and to advance the technological development in this area, not to encourage or support applications that may violate ethical standards.

\noindent\textbf{Compliance with Licensing and Distribution Regulations.} We affirm that all open-source resources utilized in our study, including detectors, language models, and datasets, have been employed in strict accordance with their respective licenses and distribution terms. This adherence extends to ensuring that any modifications, redistributions, or applications of these resources in our research comply with their original licensing agreements. Our commitment to these principles upholds the integrity of our research and contributes to a responsible and ethical academic environment.

\noindent\textbf{Use of Publicly Available Data and Consideration for Privacy.} The datasets used in our research are exclusively sourced from publicly available, open-source collections. While these datasets are publicly accessible and generally considered devoid of sensitive personal information, we acknowledge the potential for inadvertent inclusion of personal identifiers in datasets. We emphasize that our use of these datasets is aligned with their intended purpose and distribution terms. We also recognize the importance of respecting privacy and are committed to ongoing vigilance in this regard.

We reiterate that this research adheres to strict scientific and ethical standards, aiming to contribute to the field of natural language processing while ensuring that the results are not used for improper purposes. We also encourage our peers to consider these ethical factors when utilizing our research findings, ensuring their applications do not adversely affect society and individuals.

\section{Acknowledgements}
Lichao Sun is supported by the National Science Foundation Grants CRII-2246067.

\bibliography{custom}

\clearpage

\appendix

\section{Full Related Works}
\label{sec: appendix A}
\subsection{Detecting Machine Generated Text}
Current MGT detection methods can be broadly categorized into metric-based and model-based methods according to previous study \citep{he2023mgtbench}. Moreover, other detection methods such as watermark, retrieval-based methods, and in-context learning leveraging LLMs also lead to promising detection methods.

\noindent\textbf{Metric-based Methods.} 
Metric-based methods leverage the LLM backbone directly to extract its distinguishing features between HWT and MGT, operating within a white-box setting that requires access to the model. Former methods like Log-Likelihood \citep{solaiman2019release}, Entropy, Rank \cite{gehrmann2019gltr}, and Log-Rank \citep{mitchell2023detectgpt} employ statistical analysis to measure information beyond the token level. GLTR \citep{gehrmann2019gltr} utilizes a suite of metric-based methods to aid in human identification. However, with the advent of LLMs, the progressively increasing similarity between the distributions of HWT and MGT has weakened its detection accuracy \citep{ghosal2023towards}. 

Building upon the observation that MGTs occupy regions with sharp negative log probability curvature, \citet{mitchell2023detectgpt} introduced a zero-shot whitebox detection method called DetectGPT, setting a trend in metric-based detection \citep{su2023detectllm, mireshghallah2023smaller, bao2023fast}. \citet{yang2023dna} also introduced a powerful detection method known as DNA-GPT, which leverages N-gram in a black-box setting by analyzing the differences between truncated original text and regenerated text. Recently, they even extended the detection method to MGT code in a zero-shot setting, which is proven to achieve promising results \citep{yang2023zero}.

\begin{figure*}[h]
    \centering
    \label{fig:POS_distribution}
    \includegraphics[width=\linewidth]{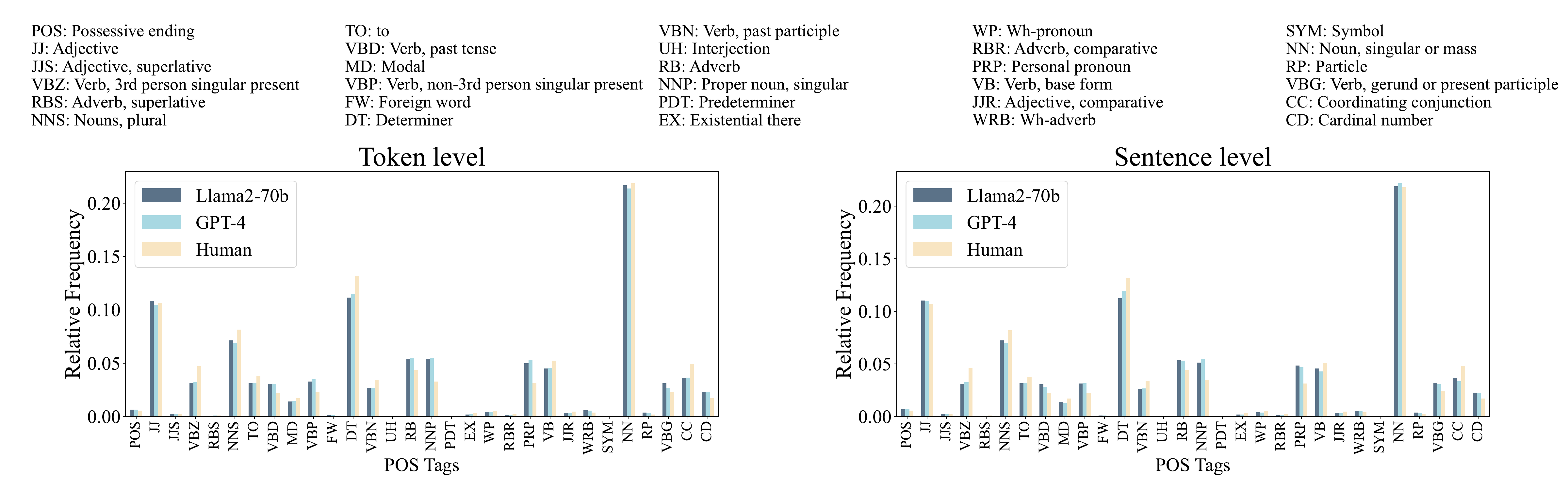}
    \caption{POS distribution of the \textsc{MixSet} by NLTK \citep{nltk2009}.}
\end{figure*}

\begin{figure*}[h]
    \centering
    \includegraphics[width=\linewidth]{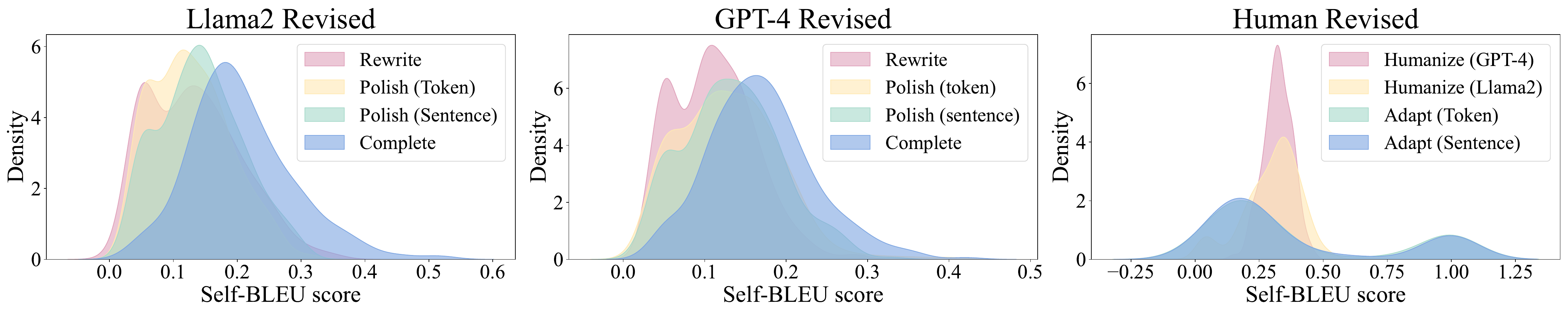}
    \caption{Self-BLEU score of the HWT, MGT, and MixSet.}
    \label{fig:self-BLEU}
\end{figure*}

\noindent\textbf{Model-based Methods.}
In the Large Language Models (LLMs) era, \citet{guo2023close} developed the ChatGPT Detector based on a fine-tuned Roberta model. As for decoder-based or encoder-decoder detectors, GPT-sentinel \citep{chen2023gpt} and RADAR \citep{hu2023radar}, utilizing T5-small \citep{raffel2020exploring} and Vicuna-7B \citep{vicuna2023} respectively, show convincing results when detecting MGT even in revised cases. Moreover, \citet{verma2023ghostbuster} proposes a novel detection framework called Ghostbuster, which employs passing documents through a series of weaker language models. Using a small amount of training data, \citet{guo2023authentigpt} leverages a black-box LLM to denoise input text with artificially added noise and then semantically compares the denoised text with the original to determine if the content is machine-generated, leading a new method for MGT detection.

However, it's important to note that some researchers have raised concerns about fine-tuning models for MGT detection. \citet{bakhtin2019real} and \citet{uchendu2020authorship}  have argued that fine-tuning models can lead to overfitting and a loss of generalization, particularly when dealing with text generated by the latest LLMs. They highlight the challenge posed by out-of-distribution editing texts, which can undermine the effectiveness of pre-trained detectors, as demonstrated by research on paraphrasing. 

\noindent\textbf{Other detection methods.}
Watermarking imprints specific patterns of the LLM output text that can be detected by an algorithm while being imperceptible to humans. \citet{kirchenbauer2023watermark} developed watermarks for language modeling by adding a green list of tokens during sampling. Currently, \citet{gu2023learnability} introduces a learnable watermark by distilling LLM and watermark technology into one student model, finding that models can learn to generate watermarked text with high detectability.

In retrieval-based methods, \citet{krishna2023paraphrasing} introduce a method to retrieve semantically similar generations and search a database of sequences previously generated by specific Large Language Models (LLMs), looking for sequences that match the candidate text within a certain threshold. Delving deeper, \citet{wu2023llmdet} proposes a model-specific detection tool called LLMDet, which can detect source text from specific LLMs by constructing a text collection dictionary for each LLM.

In the in-context learning setting, \citet{yu2023gpt} introduced a straightforward method that analyzes the similarity between re-answering a question by generating a corresponding question in the context of the given answer. Moreover, \citet{koike2023outfox} employed a pure in-context learning approach for detection and found that LLMs can distinguish between human and machine styles.

\subsection{Previous study to mix of HWT and MGT}
Prior studies have viewed the mixture of HWT and MGT in different settings. DNA-GPT \citep{yang2023dna} and DetectGPT \citep{mitchell2023detectgpt} notably utilized the T5 model \citep{raffel2020exploring} to simulate scenarios where humans make limited, random modifications to MGT, creating complex test cases. Conversely, DIPPER \citep{krishna2023paraphrasing} and OUTFOX \citep{koike2023outfox} opted for a paraphrasing technique, using this method to craft adversarial attacks aimed at eluding the detection mechanisms of classifiers, thereby presenting a nuanced way to alter MGT while maintaining undetectability.
Recent research efforts have started to explore real-world applications of human-AI mixtext. \citet{peerreview} explores the impact of AI, such as ChatGPT, on modifying content in academic peer reviews, aligning with our focus on the detection of mixtext.

\subsection{Datasets for MGT Detection}
Previous studies have proposed many datasets of MGT, often accompanied by their newly proposed detectors \citep{verma2023ghostbuster, chen2023gpt}. \citet{guo2023close} leverages multiple previous Question-Answer (QA) datasets \citep{rajpurkar2016squad, kovcisky2018narrativeqa, jin2019pubmedqa, lin2021truthfulqa}, allowing ChatGPT to generate corresponding answers without explicit prompts. This approach results in creating a comprehensive dataset comprising a large set of pairs of MGT and HWT. Following the QA pattern, many researchers \citep{mitchell2023detectgpt, su2023detectllm, hu2023radar, he2023mgtbench} propose datasets with the MGT from variant mainstream LLMs \citep{du2022glm, black2022gpt, anand2023gpt4all,  openai2022gpt35, openai2023gpt4} \footnote{https://www.databricks.com/blog/2023/04/12/dolly-first-open-commercially-viable-instruction-tuned-llm}. \citet{yu2023gpt} only utilizes the answer section within the QA dataset \citep{hamborg2017news, moller2020covid} and employs ChatGPT to generate corresponding questions and re-answers.

However, these datasets typically consist of two distinct classes of texts, namely pure MGT or HWT, without accounting for the potential mixtext. Furthermore, issues arise due to variations in prompts \citep{koike2023you}, sampling methods, and the inherent differences in length, style, and quality among texts in some datasets \cite{he2023mgtbench}. These variations challenge the generalization of proposed detectors \citep{xu2023generalization} and lie a vast diversity in distribution between the original and revised text \citep{ghosal2023towards}. In some instances, the MGT included in datasets may not undergo thorough and careful evaluation. Many noisy sentences are not filtered well in the datasets. For example, some sentences like \textit{Let me know if you have any other questions} exist in the dataset, which will impact the effectiveness of the detectors \citep{guo2023close}.

\section{Dataset Details}
\label{sec: appendix B}
\subsection{Construction Details}
\label{sec: appendix B.1}

% \begin{table}[h!]
% \renewcommand\arraystretch{1.1}
% \centering
% \caption{Detailed Distribution of Different Operations in MIXSET}
% \label{tab: MIXSET_distribution}
% \begin{tabular}{lccc}
% \toprule[1.5pt]
% \textbf{Operation} & \textbf{GPT-4} & \textbf{Llama2} & \textbf{Human} \\
% \toprule[1.5pt]
% AI Revised &    &        &        \\
% \hline
% Polish Tok.       & 300   & 300    & —      \\
% Polish Sen.       & 300   & 300    & —      \\
% Complete          & 300   & 300    & —      \\
% Rewrite           & 300   & 300    & —      \\
% \hline
% % 
% \multicolumn{2}{l}{Human Revised}  &   &     \\
% \hline
% Humanize          & —     & —      & (300+300)    \\
% Adapt Tok.        & —     & —      & 300    \\
% Adapt Sen.        & —     & —      & 300    \\
% \bottomrule[1.5pt]
% \end{tabular}
% \end{table}

\renewcommand\arraystretch{1}

\begin{figure*}
    \includegraphics[width=\textwidth]{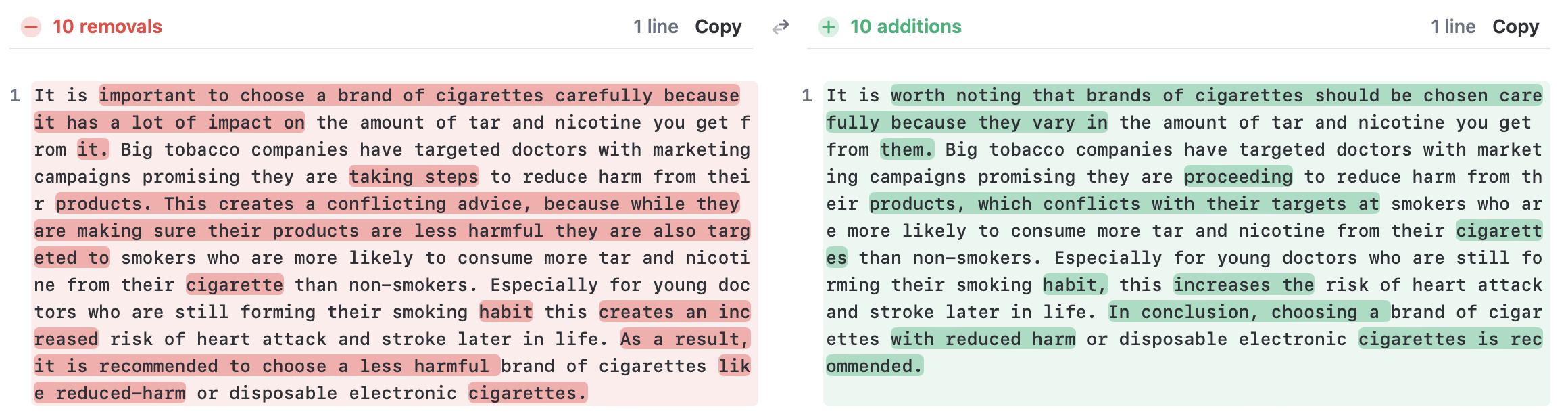}
    \caption{screenshot of human revising on MGT}
    \label{screenshot_label}
\end{figure*}

\noindent\textbf{Eight Human revised the MGT to mixtext.}
The MGT is revised by eight human experts with professional English proficiency and costs them a total of 280 hours to complete this part. The guidelines for human revision are shown in Figure \ref{Guidelines for Human Revision}. And the labeling screenshot is shown in Figure \ref{screenshot_label}.

\subsection{Other Metrics in Evaluating \textsc{MixSet}}
\label{sec: appendix B2}
\begin{figure}[ht]
    \centering
    \includegraphics[width=\linewidth]{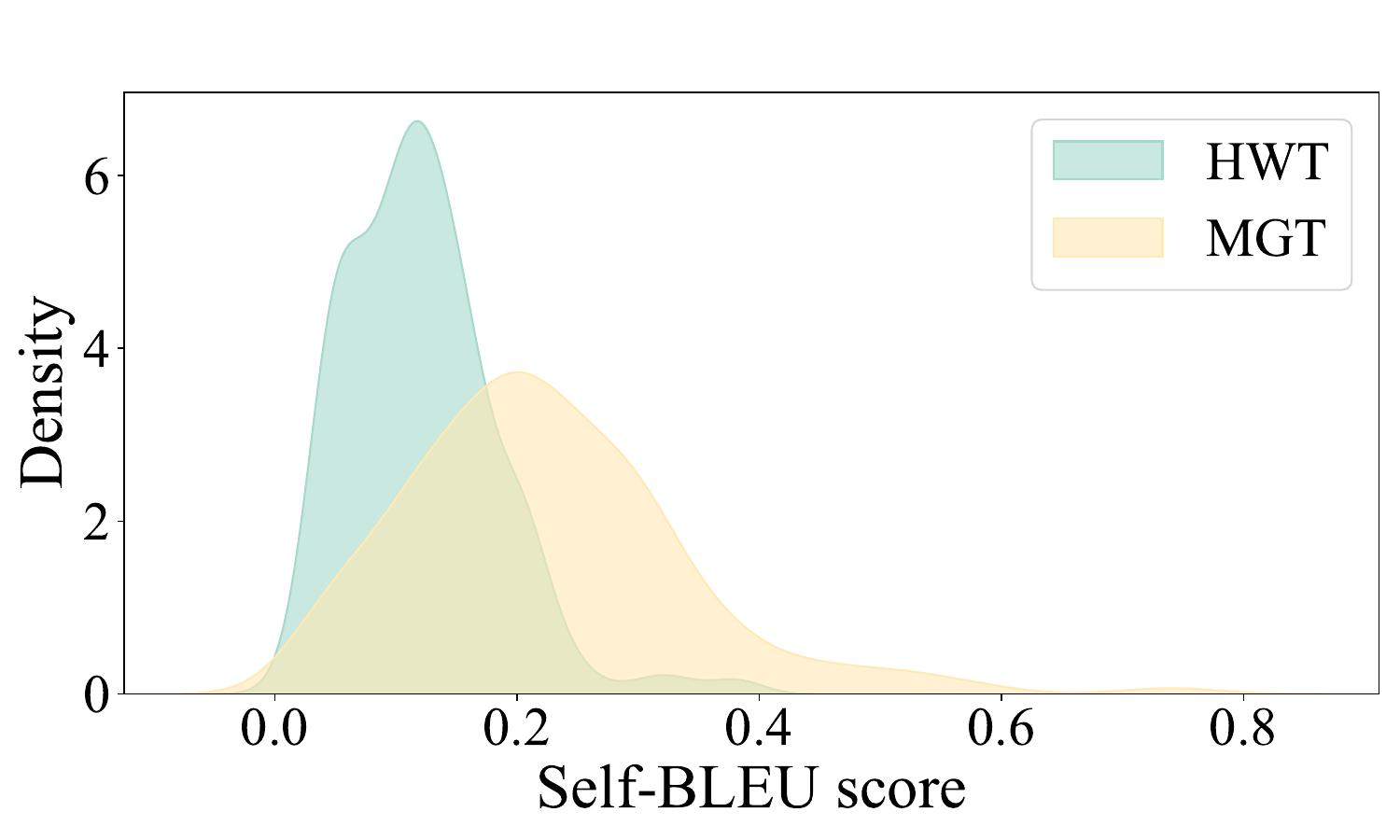}
    \caption{Self-BLEU score of HWT and MGT.}
    \label{fig:self-BLEU_raw}
\end{figure}

\begin{itemize}[nolistsep, leftmargin=*]
    \item \textbf{Self-BLEU Score: }Self-BLEU is a metric used to assess the diversity of generated text. Generally, a lower Self-BLEU score indicates higher textual diversity. These results are shown in Figure \ref{fig:self-BLEU}. Overall, the HWT shows greater diversity than MGT, and the Rewrite category has the highest textual diversity in the MixSet. The self-BLEU score of HWT, WGT, and mixtext is shown in Figure \ref{fig:self-BLEU_raw} and \ref{fig:self-BLEU}.

    \item \textbf{POS distribution: }POS distribution refers to the frequency and pattern of Part-of-Speech tags in a text, categorizing words into grammatical classes like nouns, verbs, and adjectives. This analysis is key for understanding the text's syntactic structure and linguistic characteristics, which is important in NLP research fields.
\end{itemize}

\noindent\textbf{Seven Model-Based detectors.} We implement seven Machine Generative Text (MGT) detectors, encompassing both supervised and zero-shot settings. Firstly, we consider a robust closed-source online detector baseline: GPTZero \citep{tian2023gptzero}. Secondly, we implement three open-source encoder-based detectors: OpenAI's classifier \citep{Openai2023classifier}, Roberta-based classifier \citep{guo2023close}. We also implement GPT-sentinel \citep{chen2023gpt}, RADAR \citep{hu2023radar}, and Ghostwriter \cite{verma2023ghostbuster} as strong baselines. We also finetune a pre-trained language model built by \citet{sanh2019distilbert} with an extra classification layer on top. 

\noindent\textbf{Three Evaluation Metrics} Previous studies \citep{sadasivan2023can, mitchell2023detectgpt} have proven the feasibility of using the Area Under The ROC Curve (AUROC) score for evaluating detection algorithm effectiveness. Given that most detectors can only give a predictive probability, we build a logistic regression model to provide concrete predictions, i.e., MGT or HWT, converting probability to accuracy and f1-score as the two other metrics for our detection evaluation.

\begin{figure}[H]
    \centering
    \includegraphics[width=\linewidth]{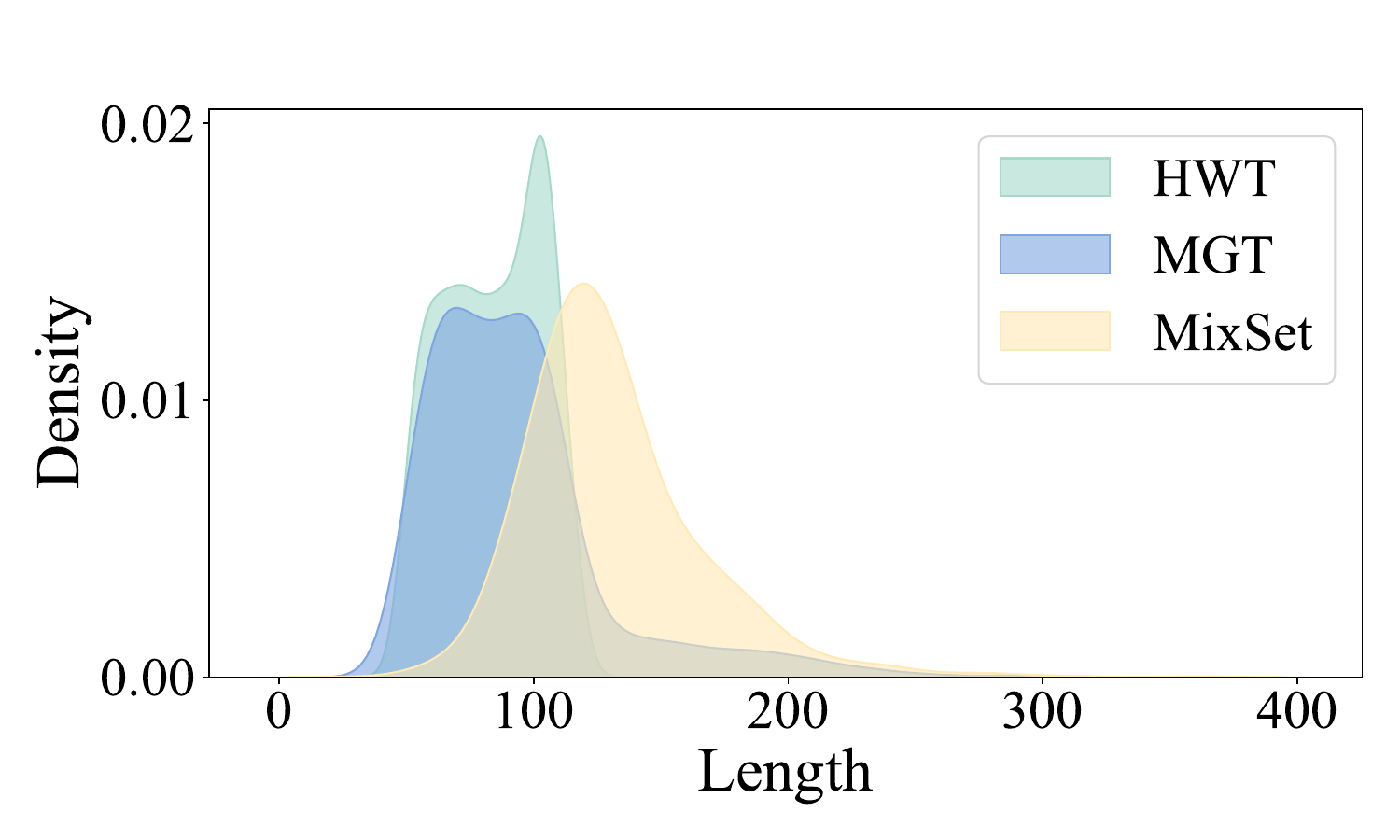}
    \caption{Length distribution of the training datasets and the MixSet.}
    \label{fig:length_training}
\end{figure}

\noindent\textbf{Training set construction.}  We respectively select pure HWT and MGT for the train set from different datasets as illustrated in \ref{HWTdataset} and MGTBench \cite{he2023mgtbench}, which is also the original dataset of our \textsc{MixSet}. Since all datasets are specific, this selection strategy ensures only a small difference in data distribution. Firstly, we do data deduplication and pre-process it to erase the Unicode or other special tokens like \textit{\texttt{\textbackslash n\textbackslash n}}. Then, we select pieces of sentences with a similar length distribution in our \textsc{MixSet}, as illustrated in Figure \ref{fig:length_training}. As we use accuracy as our evaluation metric, we restrict the amount of HWT and MGT to be the same in our dataset, as illustrated in Tabel \ref{tab: train set size}.

\noindent\textbf{Training details.}
We employ the standard binary-classification loss function and the AdamW optimizer \citep{loshchilov2019decoupled}, with an empirically determined learning rate. Specifically, for the Hello-Ai/Roberta-based model and the DistilBERT model, the learning rate is set to \(5 \times 10^{-7}\). In contrast, for Radar and GPT-sentinel, the learning rates are \(5 \times 10^{-6}\) and \(5 \times 10^{-5}\), respectively. Each supervised model undergoes training for three epochs on a dual-4090 server.

\begin{figure}[h]
    \centering
  \begin{subfigure}{0.5\textwidth}
    \includegraphics[width=\linewidth]{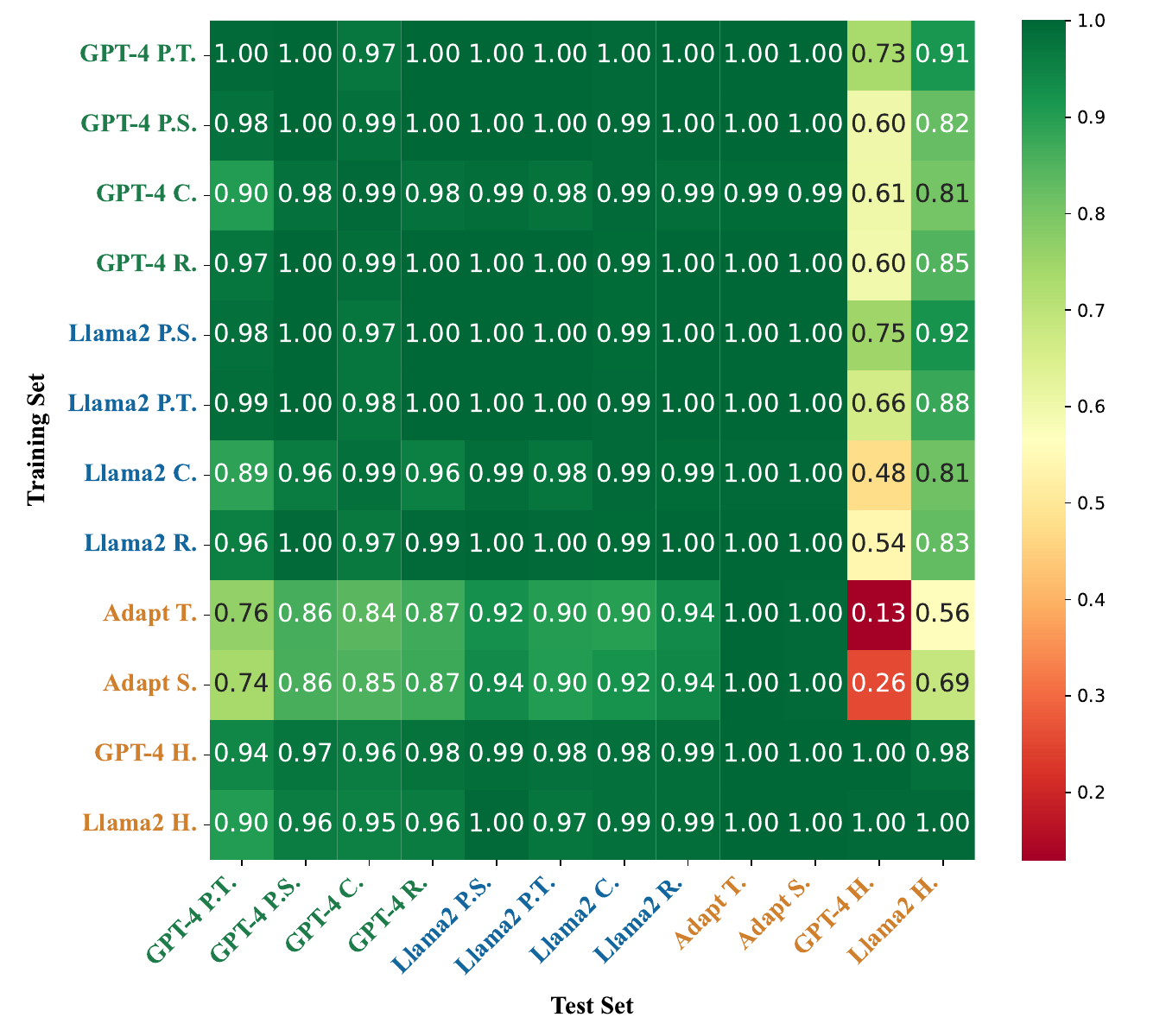}
    \caption{The AUC Heatmap of Radar}
    \label{fig:subfig1}
  \end{subfigure}
    \centering
  \begin{subfigure}{0.5\textwidth}
    \includegraphics[width=\linewidth]{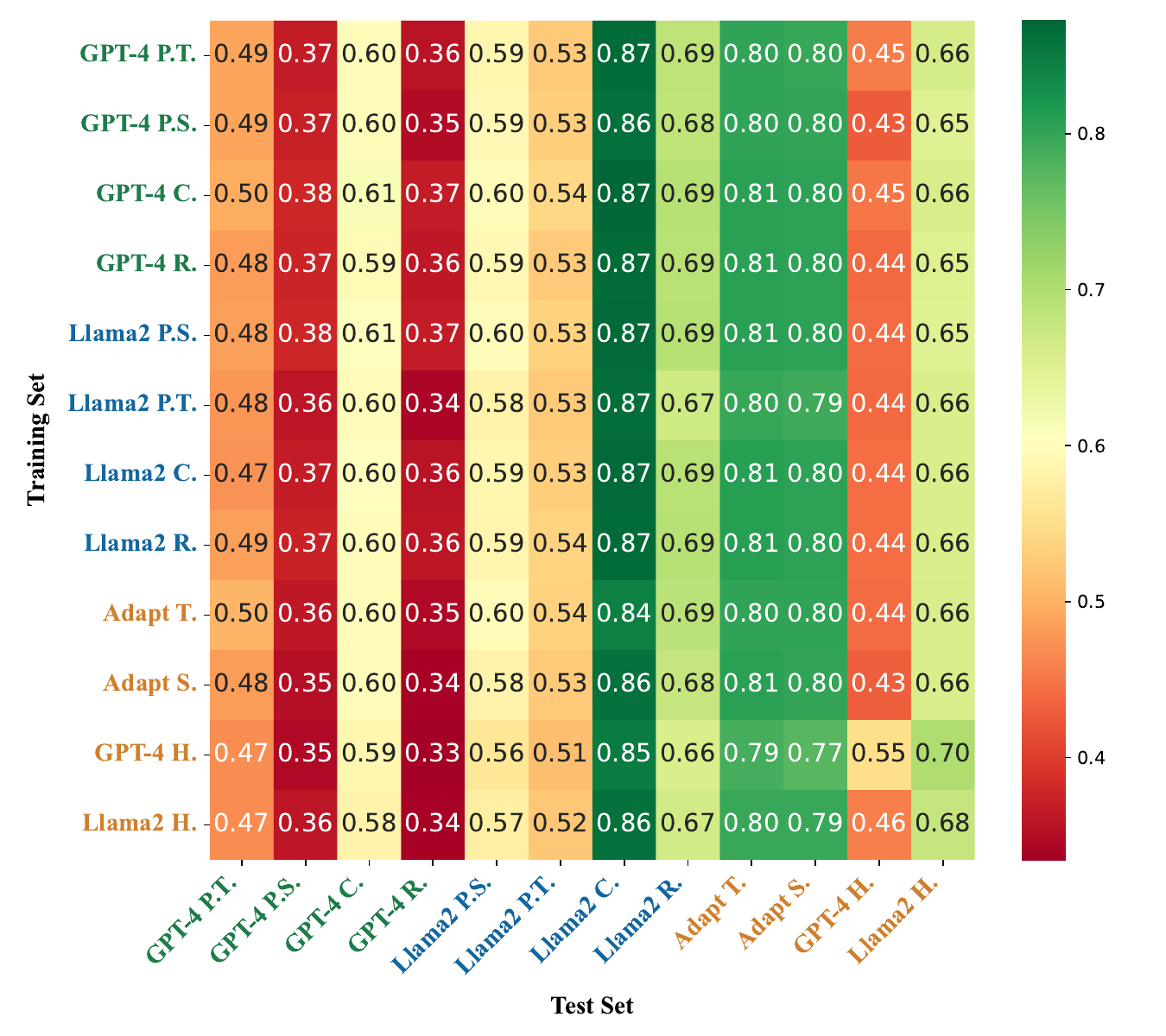}
    \caption{The AUC Heatmap of ChatGPT Detector}
    \label{fig:subfig2}
  \end{subfigure}
    \centering
  \begin{subfigure}{0.5\textwidth}
    \includegraphics[width=\linewidth]{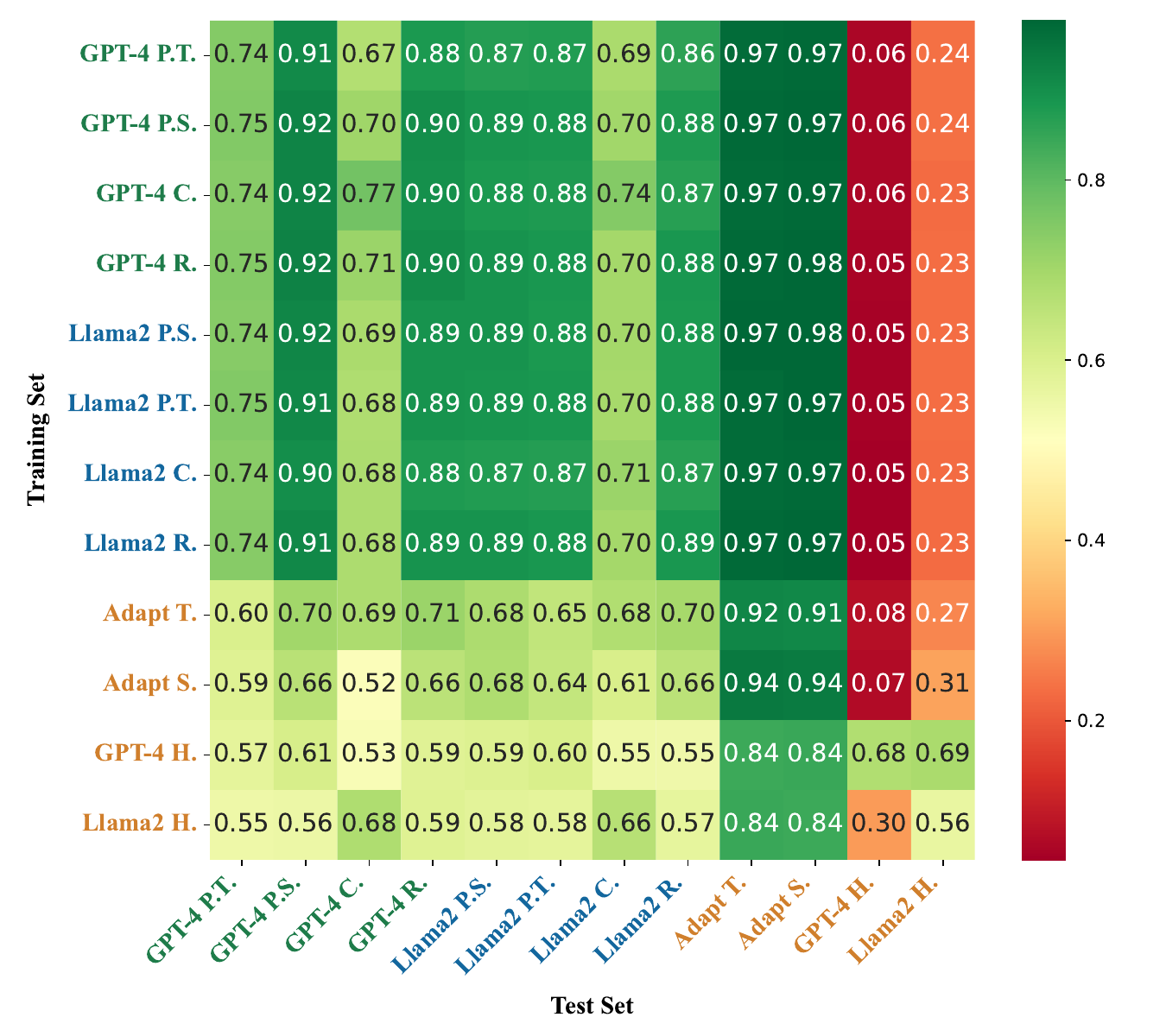}
    \caption{The AUC Heatmap of distilbert-based}
    \label{fig:subfig3}
  \end{subfigure}
  \caption{The AUC Heatmap of the other three detectors. }
  \label{fig:entire_figure}
\end{figure}

\section{Detailed Experiment Results}
\label{sec: appendix D}
\begin{figure*}
    \includegraphics[width=\textwidth]{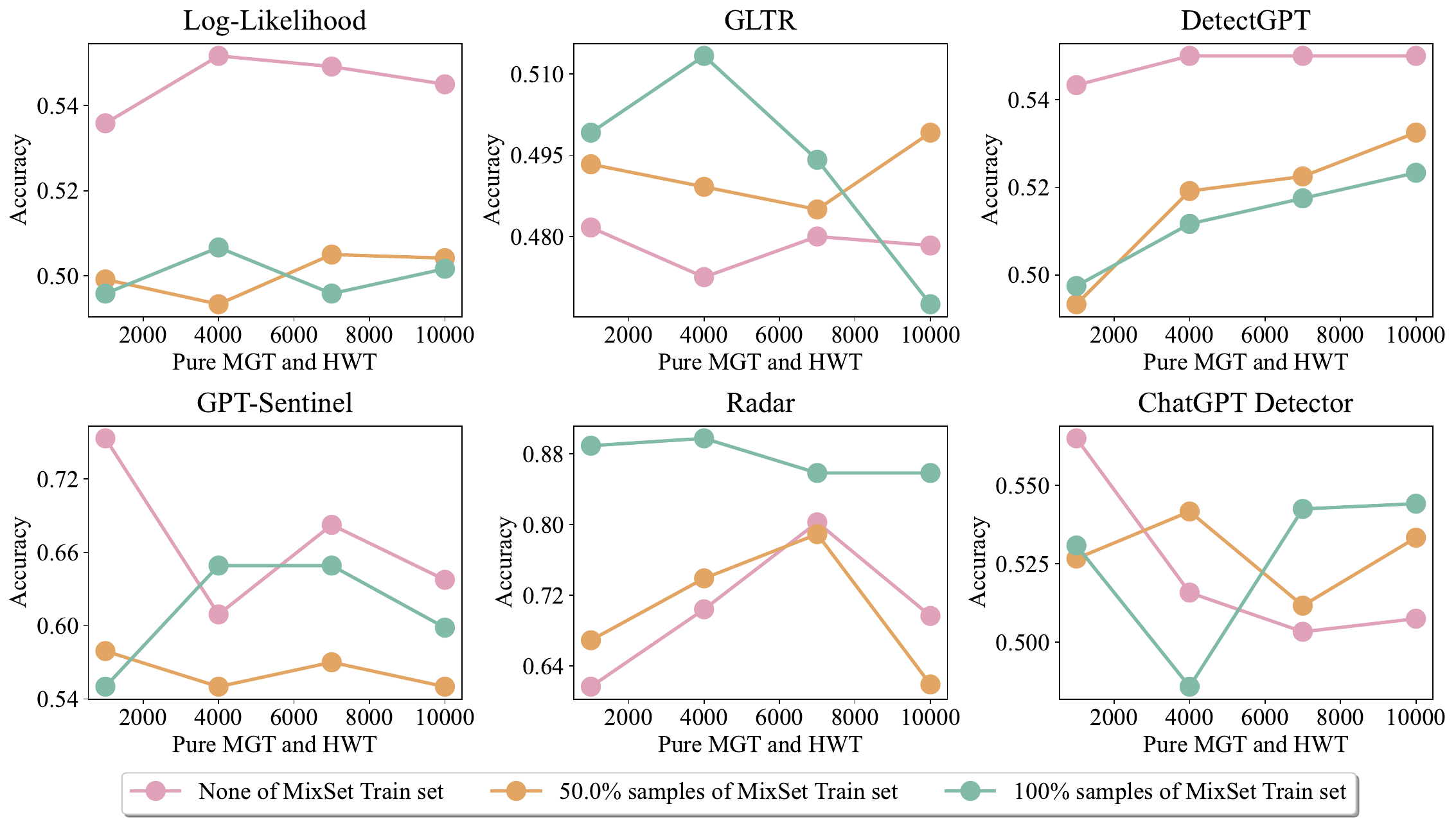}
    \caption{Analysis of the accuracy of various detectors across differing quantities of \textit{mixtext} instances from \textsc{MixSet}, as well as pure MGT and HWT.}
    \label{fig:ex4_acc}
\end{figure*}

\begin{figure*}
    \includegraphics[width=\textwidth]{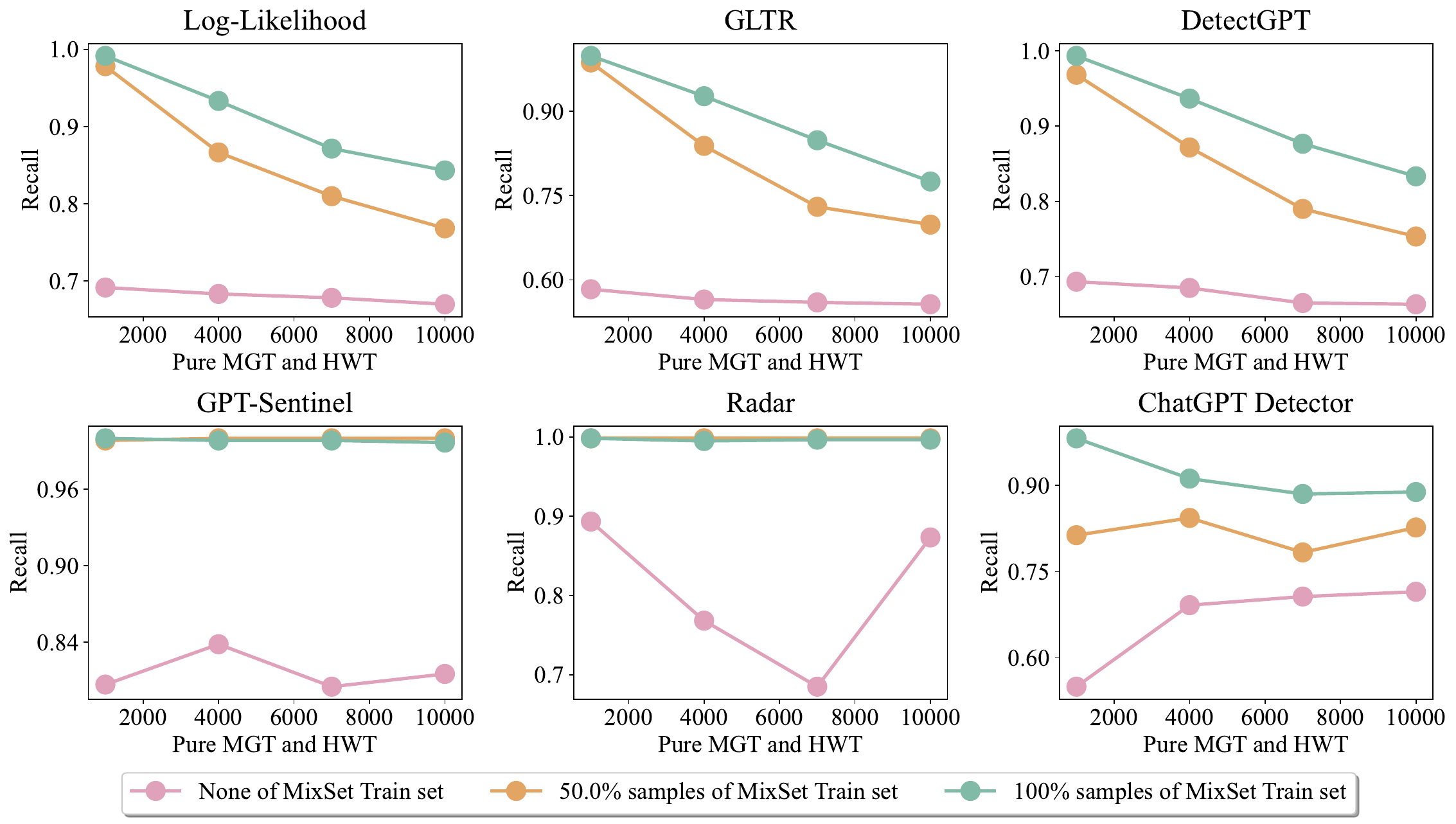}
    \caption{Analysis of the recall rate of various detectors across differing quantities of \textit{mixtext} instances from \textsc{MixSet}, as well as pure MGT and HWT.}
    \label{fig:ex4_recall}
\end{figure*}

\begin{figure*}
    \includegraphics[width=\textwidth]{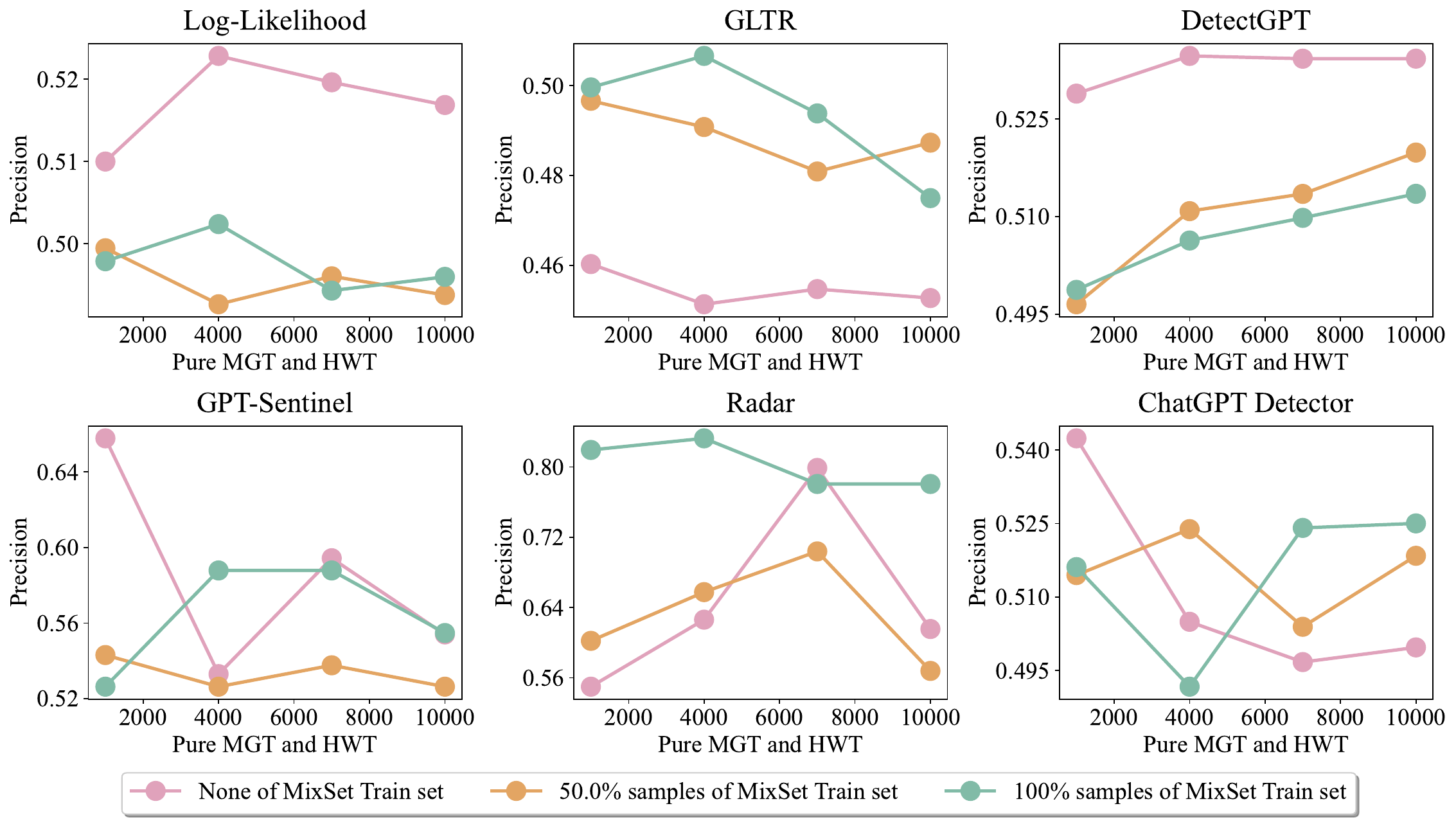}
    \caption{Analysis of the precision rate of various detectors across differing quantities of \textit{mixtext} instances from \textsc{MixSet}, as well as pure MGT and HWT.}
    \label{fig:ex4_precision}
\end{figure*}

As for experiment 1, we put the detailed accuracy for different detectors in Table \ref{tab: Ex1 details}. In experiment 2, we also evaluate detectors with AUC metric, as shown in Table \ref{tab: AUC of Ex2}. We also post other detectors undergo our experiment 3 illustrated in Figure \ref{fig:entire_figure}. As for experiment 4, we evaluate detectors with accuracy, precision, and recall metrics, as illustrated in Figure \ref{fig:ex4_acc}, \ref{fig:ex4_recall}, and \ref{fig:ex4_precision}.

\section{Prompt Template}
\label{sec: appendix E}
We show the prompt template of LLM's operation, including complete, polish (token-level and sentence-level), rewrite, and humanize in Figure \ref{completetemplate}, Figure \ref{polishtoken}, Figure \ref{polishsentence}, Figure \ref{rewritetemplate} and Figure \ref{humanizetemplate}.

\section{Case study in \textsc{MixSet}}
\label{sec: Appendix F}

We selected two cases to show the comparison between the revised mixtext and the original text, where the highlighted content represents the modified content. The HWT original text can be found in figure \ref{Case Study of revision on HWT: Original HWT}, the AI revised text are shown in Figure \ref{Case Study of revision on HWT: GPT-4 Polish (Token)}, \ref{Case Study of revision on HWT: GPT-4 Polish (Sentence)}, \ref{Case Study of revision on HWT: GPT-4 Complete}, \ref{Case Study of revision on HWT: GPT-4 Rewrite}, \ref{Case Study of revision on HWT: Llama2 Polish (Token)}, \ref{Case Study of revision on HWT: Llama2 Polish (Sentence)}, \ref{Case Study of revision on HWT: Llama2 Complete}, and \ref{Case Study of revision on HWT: Llama2 Rewrite}. The MGT original text can be found in Figure \ref{Case Study of revision on MGT: Original MGT}, and the Human revised text can be found in Figure \ref{Case Study of revision on MGT: GPT-4 Humanize}, \ref{Case Study of revision on MGT: Llama2 Humanize}, \ref{Case Study of revision on MGT: Adapt Token}, and \ref{Case Study of revision on MGT: Adapt Sentence}.

\begin{table*}[]
\renewcommand{\arraystretch}{1.5}
\small
\centering
\setlength{\tabcolsep}{6pt}
\renewcommand\arraystretch{1.3}

\caption{Percentage of identifying samples as MGT of different sets in Experiment 1. For example, the Log-Rank detector categorizes 57.30\% of samples in the Llama2-revised set as MGT. We \underline{underscore} the best-performing detector and \textbf{bold} the score greater than $0.8$, which we consider as a baseline threshold for detection. (Tok. stands for token level, and Sen. stands for sentence level)}
\scalebox{0.85}{
\begin{tabular}{l|cc|cccccccccccc}
\toprule[1.5pt]

\multicolumn{1}{c|}{\multirow{5}{*}{\textbf{Detection Method}}} & \multicolumn{1}{c}{\multirow{5}{*}{\rotatebox[origin=c]{90}{\textbf{HWT}}}} & \multicolumn{1}{c|}{\multirow{5}{*}{\rotatebox[origin=c]{90}{\textbf{MGT}}}} & \multicolumn{8}{c}{\textbf{AI-Revised}}       & \multicolumn{4}{c}{\textbf{Human-Revised}}   \\

 \cmidrule(lr){4-11} \cmidrule(lr){12-15}

\multicolumn{1}{c}{}                                  & 
\multicolumn{1}{|c}{} & \multicolumn{1}{c|}{} & \multicolumn{2}{c}{Rewrite} & \multicolumn{2}{c}{Complete}            & \multicolumn{2}{c}{Polish-Tok.} & \multicolumn{2}{c}{Polish-Sen.} & \multicolumn{2}{c}{Humanize}                              & \multicolumn{1}{c}{\multirow{2}{*}{\rotatebox[origin=c]{90}{Adapt-Tok.}}} & \multicolumn{1}{c}{\multirow{2}{*}{\rotatebox[origin=c]{90}{Adapt-Sen.}}} \\

\cmidrule(lr){4-5}  \cmidrule(lr){6-7} \cmidrule(lr){8-9} \cmidrule(lr){10-11} \cmidrule(lr){12-13} 
\multicolumn{1}{c|}{}                                  & \multicolumn{1}{c}{} & \multicolumn{1}{c|}{} & \rotatebox[origin=c]{90}{Llama2}      & \rotatebox[origin=c]{90}{GPT-4}      & \rotatebox[origin=c]{90}{Llama2} & \multicolumn{1}{c}{\rotatebox[origin=c]{90}{GPT-4}} & \rotatebox[origin=c]{90}{Llama2}        & \rotatebox[origin=c]{90}{GPT-4}        & \rotatebox[origin=c]{90}{Llama2}          & \rotatebox[origin=c]{90}{GPT-4}         & \multicolumn{1}{l}{\rotatebox[origin=c]{90}{Llama2}} & \multicolumn{1}{l}{\rotatebox[origin=c]{90}{GPT-4}} & \multicolumn{1}{c}{}                                & \multicolumn{1}{c}{}                             \\

\midrule
\multicolumn{15}{l}{Metric-based Detector} \\
\midrule

Log-rank & 0.213 & \textbf{0.847} & 0.573  & 0.240  & \textbf{0.810}  & 0.520  & 0.573  & 0.383  & 0.427  & 0.350  & 0.703  & 0.093  & 0.783  & 0.770   \\ 

Log-likelihood & 0.223 & \textbf{0.867} & 0.600  & 0.287  & \textbf{0.823}  & 0.560  & 0.643  & 0.450  & 0.513  & 0.410  & 0.703  & 0.083  & 0.790  & 0.777   \\

GLTR & 0.207 & \textbf{0.840} &0.480  & 0.180  & \textbf{0.813}  & 0.393  & 0.517  & 0.283  & 0.390  & 0.313  & 0.630  & 0.053  & 0.783  & 0.760   \\

DetectGPT & 0.350 & \textbf{0.823} & 0.643  & 0.343  & 0.743  & 0.557  & 0.650  & 0.480  & 0.563  & 0.437  & \underline{\textbf{0.807}}  & 0.533  & 0.623  & 0.597 \\

Entropy & 0.353 & \textbf{0.840}  & 0.733  & 0.580  & 0.793  & 0.623  & 0.793  & 0.730  & 0.713  & 0.640  & 0.737  & 0.223  & 0.793  & 0.770   \\ \midrule
\multicolumn{13}{l}{Model-based Detector} \\
\midrule
Openai Classifier & 0.060 & 0.607 & 0.150  & 0.047  & 0.407  & 0.037  & 0.123  & 0.037  & 0.103  & 0.053  & 0.023  & 0.007  & 0.490  & 0.453   \\

ChatGPT Detector & 0.040 & 0.757 & 0.380  & 0.157  & 0.523  & 0.287  & 0.380  & 0.130  & 0.243  & 0.117  & 0.457  & 0.097  & 0.750  & 0.770   \\

Radar & 0.307 & \textbf{0.857} & 0.730  & 0.477  & \underline{\textbf{0.893}}  & \underline{0.783}  & 0.607  & 0.447  & 0.560  & 0.387  & 0.347  & 0.037  & \textbf{0.850}  & \textbf{0.890}   \\

GPT-Sentinel & 0.133 & \textbf{0.887} & \textbf{0.833}  & \underline{\textbf{0.877}}  & 0.540  & 0.573  & \underline{\textbf{0.883}}  & \underline{\textbf{0.807}}  & 0.710  & 0.460  & 0.033  & 0.000  & \textbf{0.910}  & \textbf{0.910}   \\

Distillbert & \underline{0.483} & \underline{\textbf{0.993}} & 0.593  & 0.660  & 0.530  & 0.573  & 0.607  & 0.580  & 0.547  & 0.527  & 0.170  & 0.003  & \underline{\textbf{1.000}}  & \underline{\textbf{1.000}}   \\

Ghostbuster & 0.103 & 0.610 & \underline{\textbf{0.870}}  & 0.780  & 0.750  & 0.087  & 0.353  & 0.493  & 0.473  & 0.663  & 0.567  & \underline{0.637}  & 0.700  & 0.443   \\

GPTZero & 0.017 & 0.730 & 0.493  & 0.167  & \textbf{0.810}  & 0.177  & 0.497  & 0.260  & \underline{0.777}  & \underline{0.763}  & 0.717  & 0.187  & 0.720  & 0.497  \\

\bottomrule[1.5pt]
\end{tabular}
}
\renewcommand{\arraystretch}{1} % 重置为默认行高
\label{tab: Ex1 details}
\end{table*}

\begin{table*}[]
\renewcommand{\arraystretch}{1.5}
\small
\centering
\setlength{\tabcolsep}{6pt}
\renewcommand\arraystretch{1.3}

\caption{AUC of Experiment 2 (a). We \underline{underscore} the best-performing detector and \textbf{bold} the score greater than $0.8$, which we consider as a baseline threshold for detection. (Tok. stands for token level and Sen. stands for sentence level)}
\label{tab: AUC of Ex2}
\scalebox{0.95}{
\begin{tabular}{l|cccccccccccc}
\toprule[1.5pt]

\multicolumn{1}{c|}{\multirow{5}{*}{\textbf{Detection Method}}} & \multicolumn{8}{c}{\textbf{AI-Revised}}             & \multicolumn{4}{c}{\textbf{Human-Revised}}   \\

\cmidrule(lr){2-9} \cmidrule(lr){10-13}

\multicolumn{1}{c}{}                                  & \multicolumn{2}{|c}{Rewrite} & \multicolumn{2}{c}{Complete}            & \multicolumn{2}{c}{Polish-Tok.} & \multicolumn{2}{c}{Polish-Sen.} & \multicolumn{2}{c}{Humanize}                              & \multicolumn{1}{c}{\multirow{2}{*}{\rotatebox[origin=c]{90}{Adapt-Tok.}}} & \multicolumn{1}{c}{\multirow{2}{*}{\rotatebox[origin=c]{90}{Adapt-Sen.}}} \\

 \cmidrule(lr){2-3}  \cmidrule(lr){4-5} \cmidrule(lr){6-7} \cmidrule(lr){8-9} \cmidrule(lr){10-11} 
\multicolumn{1}{c|}{}                                  & \rotatebox[origin=c]{90}{Llama2}      & \rotatebox[origin=c]{90}{GPT-4}      & \rotatebox[origin=c]{90}{Llama2} & \multicolumn{1}{c}{\rotatebox[origin=c]{90}{GPT-4}} & \rotatebox[origin=c]{90}{Llama2}        & \rotatebox[origin=c]{90}{GPT-4}        & \rotatebox[origin=c]{90}{Llama2}          & \rotatebox[origin=c]{90}{GPT-4}         & \multicolumn{1}{l}{\rotatebox[origin=c]{90}{Llama2}} & \multicolumn{1}{l}{\rotatebox[origin=c]{90}{GPT-4}} & \multicolumn{1}{c}{}                                & \multicolumn{1}{c}{}                             \\

\midrule
\multicolumn{13}{l}{Metric-based Detector} \\
\midrule

log-rank & \textbf{0.921}  & 0.629  & 0.632  & 0.318  & 0.569  & 0.531  & 0.662  & 0.462  & 0.641  & 0.245  & 0.778  & 0.778   \\ 

log likelihood & \textbf{0.933}  & 0.650  & 0.672  & 0.352  & 0.610  & 0.569  & 0.709  & 0.508  & 0.652  & 0.206  & 0.782  & 0.786   \\ 

GLTR & \textbf{0.870}  & 0.504  & 0.546  & 0.268  & 0.511  & 0.466  & 0.602  & 0.345  & 0.595  & 0.208  & 0.764  & 0.768   \\

DetectGPT & \textbf{0.852}  & 0.644  & 0.669  & 0.352  & 0.612  & 0.466  & 0.664  & 0.482  & 0.677  & 0.461  & 0.548  & 0.557   \\

Entropy & \textbf{0.814}  & 0.581  & 0.662  & 0.463  & 0.656  & 0.636  & 0.686  & 0.596  & 0.580  & 0.185  & 0.733  & 0.730  \\ \midrule

\multicolumn{13}{l}{Model-based Detector} \\
\midrule
Openai Classifier & 0.294  & 0.601  & 0.126  & 0.360  & 0.433  & 0.492  & 0.383  & 0.590  & 0.321  & 0.517  & 0.182  & 0.187   \\

ChatGPT Detector & 0.706  & 0.399  & \textbf{0.874}  & 0.640  & 0.567  & 0.508  & 0.617  & 0.410  & 0.679  & 0.483  &\textbf{0.818}  & \textbf{0.813}   \\

Radar & \textbf{0.992}  & \underline{\textbf{0.994}}  & \underline{\textbf{0.997}}  & \underline{\textbf{0.999}}  & \underline{\textbf{0.998}}  & \underline{\textbf{0.986}}  & \underline{\textbf{0.998}}  & \underline{\textbf{1.000}}  & \underline{\textbf{0.984}}  & \underline{\textbf{0.984}} & \underline{\textbf{0.999}}  & \underline{\textbf{0.999}} \\

GPT-sentinel & \underline{\textbf{0.994}}  & \textbf{0.992}  & \textbf{0.987}  & \textbf{0.993}  & \textbf{0.995}  & \textbf{0.964}  & \textbf{0.992}  & \textbf{0.996}  & \textbf{0.915}  & \textbf{0.953}  & \textbf{0.958}  & \textbf{0.986}   \\

Distillbert & 0.756  & \textbf{0.856}  & 0.746  & \textbf{0.859}  & 0.790  & 0.730  & 0.791  & \textbf{0.856}  & 0.416  & 0.330  & \textbf{0.837}  & \textbf{0.861}   \\

\bottomrule[1.5pt]
\end{tabular}
}
\renewcommand{\arraystretch}{1} % 重置为默认行高
\end{table*}

% \clearpage

% \section{Prompt Template and Case study in \textsc{Mixset}}

% (\ding{172})

\begin{figure*}
\centering
\caption{Prompt(\ding{172})-LLM complete the HWT}

\begin{tcolorbox}[colback=gray!5!white,colframe=gray!50!black,   colbacktitle=gray!75!black]

\texttt{I have an incomplete text and need it to be completed. Please expand this into a complete text where the total word count, including the original text I have provided, does not exceed 180 words. The original text must remain exactly as is, with its format (such as capitalization and punctuation) intact. Please do not modify any part of the original text. Here's the text: \{HWT\}}

\end{tcolorbox}
\label{completetemplate}
\end{figure*}

\begin{figure*}
\centering

\caption{Prompt(\ding{173})-LLM polish HWT in token level}
\begin{tcolorbox}[colback=gray!5!white,colframe=gray!50!black,   colbacktitle=gray!75!black]

\texttt{Please carefully examine the following paragraph solely for spelling and grammatical errors, and replace any words that are repetitive, inaccurate, or poorly chosen. It is crucial to avoid any changes to the sentence order or structure. The focus should be strictly on the choice and usage of individual words to improve the clarity and appropriateness of the text without altering the original sentence construction: \{HWT\}}

\end{tcolorbox}
\label{polishtoken}
\end{figure*}

\begin{figure*}
\centering

\caption{Prompt(\ding{174})-LLM polish HWT in sentence level}
\begin{tcolorbox}[colback=gray!5!white,colframe=gray!50!black,   colbacktitle=gray!75!black]

\texttt{Please optimize the sentences in the following paragraph to enhance fluency and clarity. Do not alter the overall content or structure of the paragraph. Focus on the construction and expression of the sentences, ensuring that the text is coherent and the information is accurate: \{HWT\}}

\end{tcolorbox}
\label{polishsentence}
\end{figure*}

\begin{figure*}
\centering

\caption{Prompt(\ding{175})-LLM rewrite HWT}
\begin{tcolorbox}[colback=gray!5!white,colframe=gray!50!black,   colbacktitle=gray!75!black]

\texttt{Please extract the core ideas and keywords from the following English text and then rewrite a passage based on this information. The new text should maintain the essence of the original, with the word count varying by no more than 10\% from the original. There's no need to list the core ideas and keywords. Here is the text that needs to be processed: \{HWT\}}

\end{tcolorbox}
\label{rewritetemplate}
\end{figure*}

\begin{figure*}
\centering
\caption{Prompt(\ding{175})-LLM humanize MGT}
\begin{tcolorbox}[colback=gray!5!white,colframe=gray!50!black,   colbacktitle=gray!75!black]
%,title=Prompt(\ding{175})-LLM humanize MGT
\texttt{I need to modify a machine-generated text to make it appear more like it was written by a human. The objective is to introduce elements commonly found in human-written texts.  Here are some optional modifications you can choose to apply:}

\texttt{1. Introduce spelling errors or typos(optional) } 
       
\texttt{2. Create grammatical errors, such as randomly adding or deleting words (optional). }   

\texttt{3. Include relevant but internet links, like blog posts or image links pertaining to the topic, you don't have to use the real links, so you can freely write one (optional). }  

\texttt{4. Add relevant hashtags, for instance, \#TopicKeyword \#Location \#Activity (optional).}  

\texttt{5. Use internet slang and abbreviations, e.g., `OMG', `How r u', `LOL', (optional).}

\texttt{Please select any combination of these modifications to enhance the text's human-like quality. The aim is to simulate the imperfections and stylistic choices typical in casual human writing.} 

\texttt{The word count of the new text should not exceed 1.1 times that of the original text.} 

\texttt{You should just give me the revised version without any other words.} 

\texttt{Emojis are strictly prohibitive, so please ensure that it contains no emojis.} 

\texttt{Here is the machine-generated text:\{HWT\}}

\end{tcolorbox}
\label{humanizetemplate}
\end{figure*}

\begin{figure*}
\centering
\caption{Guidelines for Human Revision}
\label{Guidelines for Human Revision}
\begin{tcolorbox}[colback=orange!5!white,colframe=orange!50!black,   colbacktitle=orange!75!black]
%,title=Guidelines for Human Revision

\texttt{The content under this document is generated by a large language model, such as ChatGPT. You are required to revise it to make it closer to the style of human-written text. You are responsible for the text under the IDs xx-xx, and you need to make the following three types of modifications to the content, generating two different sentences for each ID (each ID corresponds to 3 sentences):}

\texttt{The document is in JSON format. You can choose to use code editors like Visual Studio Code or text editors like Notepad for reading and writing.}

\texttt{1. Adapt Token: You need to modify any words or phrases in this passage that you think are too rigid, mechanical, obscure, or unusual into vocabulary typical of human texts. Be careful not to alter the sentence order or structure; only modify `words' or `phrases.'}

\texttt{2. Adapt Sentence: You need to revise any sentence structures that you find too mechanical or rigid to make them more in line with how humans typically write texts. This involves changes at the sentence level, which may include altering sentence order and structure.}

\texttt{Note:}

\texttt{1. After writing, please ensure to check that there are no `grammatical errors' or `spelling mistakes' in the text paragraphs.}

\texttt{2. Do not use ChatGPT or other large language models for data annotation, as it will severely degrade the data quality.}

\texttt{3. You may use translation platforms like Youdao \cite{youdao_translate} or Google \footnote{\url{https://translate.google.com/}}; or use Grammarly \footnote{https://www.grammarly.com/} to check for grammatical errors.}

\texttt{Below are some examples you can refer to:}

\texttt{[Example 1]}

\texttt{[Example 2]}

\texttt{[Example 3]}

\end{tcolorbox}
\end{figure*}

%Original HWT
\begin{figure*}
\centering
\caption{Case Study of revision on HWT: Original HWT}
\label{Case Study of revision on HWT: Original HWT}
\begin{tcolorbox}[colback=green!5!white,colframe=green!50!black,   colbacktitle=green!75!black]
%,title=Case Study of revision on HWT: Original HWT

\texttt{\textbf{[Original HWT]} We present a new mixed variable symplectic (mvs) integrator for planetary systems, that fully resolve close encounters. The method is based on a time regularisation that allows keeping the stability properties of the symplectic integrators, while also reducing the effective step size whenever two planets encounter. We use a high order mvs scheme such that it is possible to integrate with large time steps far away from close encounters. We show that this algorithm is able to resolve almost exact collisions (i.e. with a mutual separation of a fraction of the physical radius) while using the same time-step as in weakly perturbed problems such as the solar system. We demonstrate the long-term behavior on systems of six super-earths experiencing strong scattering for 50 kyr. We compare our algorithm to hybrid methods such as mercury and show that for an equivalent cost, we obtain much better energy conservation.} 

\end{tcolorbox}
\end{figure*}

%GPT-4 Polish Token
\begin{figure*}
\centering
\caption{Case Study of revision on HWT: GPT-4 Polish (Token)}
\label{Case Study of revision on HWT: GPT-4 Polish (Token)}
\begin{tcolorbox}[colback=green!5!white,colframe=green!50!black,   colbacktitle=green!75!black]
%title=Case Study of revision on HWT: GPT-4 Polish (Token)

    \texttt{\textbf{[GPT-4 Polish (Token)]} We present a new \colorbox{yellow}{mixed-variable} symplectic \colorbox{yellow}{(MVS)} integrator for planetary systems that fully \colorbox{yellow}{resolves} close encounters. The method is \colorbox{yellow}{grounded in a temporal}\colorbox{yellow}{regularization that maintains} the stability properties of symplectic integrators while also \colorbox{yellow}{diminishing} the effective step size whenever two planets \colorbox{yellow}{draw near each other. We employ a high-order MVS scheme, enabling integration} with \colorbox{yellow}{extended} time steps \colorbox{yellow}{when distant} from close encounters. We \colorbox{yellow}{demonstrate} that this algorithm \colorbox{yellow}{capably resolves nearly} exact collisions (i.e., with a mutual separation of a small fraction of the physical radius) \colorbox{yellow}{using a consistent} time-step as in weakly perturbed \colorbox{yellow}{problems} such as the solar system. We \colorbox{yellow}{exhibit} the \colorbox{yellow}{long-term dynamics} on systems of six super-Earths \colorbox{yellow}{undergoing intense} scattering \colorbox{yellow}{over} 50 kyr. We \colorbox{yellow}{contrast} our algorithm \colorbox{yellow}{with} hybrid methods such as \colorbox{yellow}{Mercury, revealing that, for a comparable cost,} we \colorbox{yellow}{achieve} much \colorbox{yellow}{improved} energy conservation.}

\end{tcolorbox}
\end{figure*}

%GPT-4 Polish Sentence
\begin{figure*}
\centering
\caption{Case Study of revision on HWT: GPT-4 Polish (Sentence)}
\label{Case Study of revision on HWT: GPT-4 Polish (Sentence)}
\begin{tcolorbox}[colback=green!5!white,colframe=green!50!black,   colbacktitle=green!75!black]
%,title=Case Study of revision on HWT: GPT-4 Polish (Sentence)

\texttt{\textbf{[GPT-4 Polish (Sentence)]} We \colorbox{yellow}{introduce} a \colorbox{yellow}{novel} mixed variable symplectic \colorbox{yellow}{(MVS)} integrator \colorbox{yellow}{specifically designed} for planetary \colorbox{yellow}{systems} that \colorbox{yellow}{efficiently manages} close encounters. \colorbox{yellow}{This} method \colorbox{yellow}{utilizes time regularization to maintain the stability characteristics} \colorbox{yellow}{inherent} \colorbox{yellow}{to}symplectic integrators, while \colorbox{yellow}{dynamically adjusting} the effective step size \colorbox{yellow}{during planetary encounters. Our high-order MVS approach enables} \colorbox{yellow}{integration at large time steps when planets are distantly separated, thereby} \colorbox{yellow}{optimizing computational efficiency. The algorithm proves capable of resolving}  \colorbox{yellow}{near-exact collisions—with mutual separations as minute as a fraction of the}  \colorbox{yellow}{planets' physical radii—without altering the time step used for more stable} \colorbox{yellow}{configurations,} such as \colorbox{yellow}{within} the solar system. We \colorbox{yellow}{examine} the \colorbox{yellow}{algorithm's long-term stability in simulations of six super-Earths undergoing} \colorbox{yellow}{significant scattering over 50,000 years. Furthermore, we benchmark} our algorithm \colorbox{yellow}{against} hybrid methods like Mercury, demonstrating superior energy conservation at a comparable computational cost.}

\end{tcolorbox}
\end{figure*}

%GPT-4 Complete
\begin{figure*}
\centering
\caption{Case Study of revision on HWT: GPT-4 Complete}
\label{Case Study of revision on HWT: GPT-4 Complete}
\begin{tcolorbox}[colback=green!5!white,colframe=green!50!black,   colbacktitle=green!75!black]
%,title=Case Study of revision on HWT: GPT-4 Complete

\begin{minipage}{\linewidth}

  \texttt{\textbf{[GPT-4 Complete]} {We present a new mixed variable symplectic (mvs) integrator for planetary systems, that fully resolve close encounters. The method is based on a time regularisation that allows keeping the stability properties of the symplectic integrators, while also reducing the effective step size whenever two planets encounter. We use a \colorbox{yellow}{\parbox{\dimexpr\linewidth-2\fboxsep}{modified leapfrog scheme to address the issues of traditional integrators when handling near-collision scenarios. The key improvement lies in an adaptive algorithm that dynamically adjusts the step size according to the proximity of the planetary bodies. As a result, our integrator ensures high accuracy and energy conservation even during complex dynamical interactions. Tests demonstrate that the mvs integrator significantly outperforms standard fixed-step symplectic methods in long-term stability and computational efficiency. This advancement opens up new possibilities for accurately simulating the evolution of planetary systems over extended timescales, providing valuable insights into the dynamical behavior of multi-body systems }} \colorbox{yellow}{and enhancing our understanding of celestial mechanics.}
  }}
  
\end{minipage}

\end{tcolorbox}
\end{figure*}

%GPT-4 Rewrite
\begin{figure*}
\centering
\caption{Case Study of revision on HWT: GPT-4 Rewrite}
\label{Case Study of revision on HWT: GPT-4 Rewrite}
\begin{tcolorbox}[colback=green!5!white,colframe=green!50!black,   colbacktitle=green!75!black]
%,title=Case Study of revision on HWT: GPT-4 Rewrite
\begin{minipage}{\linewidth}

\texttt{\textbf{[GPT-4 Rewrite]} We \colorbox{yellow}{introduce} an \colorbox{yellow}{advanced} mixed variable symplectic (mvs) \colorbox{yellow}{tailored} for planetary systems, adept at handling close encounters precisely. This \colorbox{yellow}{technique} employs time regularization to \colorbox{yellow}{maintain} the stability characteristic of symplectic integrators and intelligently reduces the effective step size during \colorbox{yellow}{planetary} encounters. Employing a sophisticated mvs algorithm, it \colorbox{yellow}{facilitates} integration over large time steps when planets are not closely \colorbox{yellow}{interacting}.\colorbox{yellow}{Our findings illustrate that the algorithm effectively manages} \colorbox{yellow}{near-collision events}, simulating scenarios with planet \colorbox{yellow}{separations} just fractions of \colorbox{yellow}{their} physical sizes, using consistent time steps applicable to less \colorbox{yellow}{disturbed} systems like the solar system. \colorbox{yellow}{\parbox{\dimexpr\linewidth-2\fboxsep}{The integrator's performance excels in extended simulations, such as tracking six super-earths in a dense scattering scenario for 50,000 years, showcasing superior energy conservation when contrasted with hybrid algorithms like the mercury method, and doing so computational cost.}}}
  
\end{minipage}

\end{tcolorbox}
\end{figure*}

%Llama2 Polish Token
\begin{figure*}
\centering
\caption{Case Study of revision on HWT: Llama2 Polish (Token)}
\label{Case Study of revision on HWT: Llama2 Polish (Token)}
\begin{tcolorbox}[colback=green!5!white,colframe=green!50!black,   colbacktitle=green!75!black]
%,title=Case Study of revision on HWT: Llama2 Polish (Token)

\texttt{\textbf{[Llama2 Polish (Token)]} We \colorbox{yellow}{introduce} a \colorbox{yellow}{novel} mixed variable symplectic (MVS) integrator for planetary systems that \colorbox{yellow}{accurately resolves} close encounters. \colorbox{yellow}{Our} method is \colorbox{yellow}{built upon} a time regularization that \colorbox{yellow}{preserves} the stability \colorbox{yellow}{attributes} of symplectic integrators while \colorbox{yellow}{adaptively decreasing} the effective step size during \colorbox{yellow}{planetary interactions}. By \colorbox{yellow}{employing} a high-order MVS scheme, we can integrate with large time steps in regions where close encounters are not \colorbox{yellow}{occurring}. This algorithm \colorbox{yellow}{enables} the precise resolution of \colorbox{yellow}{nearly} exact collisions (with a mutual separation of a small fraction of the physical radius) using the same time step as in weakly perturbed problems, such as the solar system. We demonstrate the long-term behavior of our algorithm on systems \colorbox{yellow}{consisting of} six super-Earths \colorbox{yellow}{undergoing} intense scattering. Our approach shows superior energy conservation compared to hybrid methods like Mercury, while \colorbox{yellow}{maintaining} comparable \colorbox{yellow}{computational efficiency}.}

\end{tcolorbox}
\end{figure*}

%Llama2 Polish Sentence
\begin{figure*}
\centering
\caption{Case Study of revision on HWT: Llama2 Polish (Sentence)}
\label{Case Study of revision on HWT: Llama2 Polish (Sentence)}
\begin{tcolorbox}[colback=green!5!white,colframe=green!50!black,   colbacktitle=green!75!black]
%,title=Case Study of revision on HWT: Llama2 Polish (Sentence)

\texttt{\textbf{[Llama2 Polish (Sentence)]} We \colorbox{yellow}{introduce} a \colorbox{yellow}{novel} mixed variable symplectic (\colorbox{yellow}{MVS}) integrator for planetary systems that \colorbox{yellow}{accurately resolves} close encounters. \colorbox{yellow}{Our} method is founded on a time \colorbox{yellow}{regularization technique} \colorbox{yellow}{that} preserves the stability \colorbox{yellow}{attributes} of symplectic integrators while \colorbox{yellow}{adaptively decreasing} the effective step size \colorbox{yellow}{during planetary encounters.} \colorbox{yellow}{By employing a high-order MVS scheme,} we can integrate with substantial time steps \colorbox{yellow}{in regions unaffected by close encounters}. \colorbox{yellow}{Our algorithm} \colorbox{yellow}{efficiently handles} \colorbox{yellow}{near-collisions}(with a mutual separation of \colorbox{yellow}{a few percent} of the physical radius) \colorbox{yellow}{utilizing} the same time-step as in weakly perturbed \colorbox{yellow}{scenarios}, such as the solar system. \colorbox{yellow}{In a comprehensive analysis, we demonstrate the long-term behavior} \colorbox{yellow}{of our method} on systems \colorbox{yellow}{consisting} of \colorbox{yellow}{six super-Earths undergoing intense} scattering \colorbox{yellow}{over} 50 kyr. \colorbox{yellow}{Our findings indicate that, compared to} \colorbox{yellow}{hybrid approaches like Mercury, our algorithm achieves superior energy.} \colorbox{yellow}{conservation at an equal computational cost}}

\end{tcolorbox}
\end{figure*}

%Llama2 Complete
\begin{figure*}
\centering
\caption{Case Study of revision on HWT: Llama2 Complete}
\label{Case Study of revision on HWT: Llama2 Complete}
\begin{tcolorbox}[colback=green!5!white,colframe=green!50!black,   colbacktitle=green!75!black]
%,title=Case Study of revision on HWT: Llama2 Complete

\texttt{\textbf{[Llama2 Complete]} We present a new mixed variable symplectic (MVS) integrator for planetary systems, that fully resolves close encounters. The method is based on a time regularization that allows keeping the stability properties of the symplectic integrators, while also reducing the effective step size whenever two planets encounter. We use a \colorbox{yellow}{\parbox{\dimexpr\linewidth-2\fboxsep}{simple and efficient algorithm to detect close encounters, and switch to a smaller step size during these events to ensure accurate trajectory propagation. This approach enables us to maintain the long-term accuracy of the simulation while significantly improving computational efficiency. Our MVS integrator can handle both collisional and non-collisional close encounters, and we demonstrate its performance through several test cases involving Earth-Moon and Jupiter-Io system simulations.}}}

\end{tcolorbox}
\end{figure*}

%Llama2 Rewrite
\begin{figure*}
\centering
\caption{Case Study of revision on HWT: Llama2 Rewrite}
\label{Case Study of revision on HWT: Llama2 Rewrite}
\begin{tcolorbox}[colback=green!5!white,colframe=green!50!black,   colbacktitle=green!75!black]
%,title=Case Study of revision on HWT: Llama2 Rewrite

\texttt{\textbf{[Llama2 Rewrite]} We \colorbox{yellow}{introduce} a \colorbox{yellow}{novel} mixed variable symplectic (\colorbox{yellow}{MVS}) integrator for \colorbox{yellow}{solving} planetary \colorbox{yellow}{systems} that \colorbox{yellow}{undergo} close encounters. \colorbox{yellow}{\parbox{\dimexpr\linewidth-2\fboxsep}{Our approach combines the benefits of symplectic integrators - stability and accuracy - with the ability to adaptively adjust the time step size during close encounters. This enables us to capture near-collisions with high precision while still utilizing large time steps in stable regions. We validate our algorithm's performance by simulating systems of six super-Earths over 50,000 years, demonstrating its superiority compared to hybrid methods like Mercury in terms of energy conservation, while maintaining comparable computational}} \colorbox{yellow}{efficiency.}}

\end{tcolorbox}
\end{figure*}

%Original MGT
\begin{figure*}
\centering
\caption{Case Study of revision on MGT: Original MGT}
\label{Case Study of revision on MGT: Original MGT}
\begin{tcolorbox}[colback=blue!5!white,colframe=blue!50!black,   colbacktitle=blue!75!black]
%,title=Case Study of revision on MGT: Original MGT

\texttt{\textbf{[Original MGT]} The first coffee was brewed by the ancient egyptians over 4,000 years ago. As of 2023, coffee consumption increased 70-fold in the past 100 years. Today, 98\% of americans drink coffee. Studies have shown that drinking two cups of coffee per day is beneficial for adults. However, more and more studies also show that drinking one or less coffee is also equally beneficial. Drinking coffee has also been linked to higher memory, alertness, and focus. Overall, coffee is good for your health. }

\end{tcolorbox}
\end{figure*}

%GPT-4 Humanize
\begin{figure*}
\centering
\caption{Case Study of revision on MGT: GPT-4 Humanize}
\label{Case Study of revision on MGT: GPT-4 Humanize}
\begin{tcolorbox}[colback=blue!5!white,colframe=blue!50!black,   colbacktitle=blue!75!black]
%,title=Case Study of revision on MGT: GPT-4 Humanize

\texttt{\textbf{[GPT-4 Humanize]} \colorbox{yellow}{OMG did you kno that, like, the first cuppa joe was whipped } {\colorbox{yellow}{\parbox{\dimexpr\linewidth-2\fboxsep}{ up by ancient Egyptions like more than 4,000 yrs ago? Fast fwd to 2023, and whoa, coffee drinking's shot up a whoppin' 70 times in the last century! Nowadays, it's like 98\% of Americans are sippin' on coffee. Some smarty-pants studies point out that knockin' back two cups a day's pretty good for grown-ups. But hold up, more research is spillin' the beans that maybe just one cup or even less can do the trick too. Crazy, right? Srsly tho, chuggin' coffee's got a rep for boostin' your memory and makin' ya more alert and focused. All in all, java's a thumbs up for your health.}}}}

\texttt{\colorbox{yellow}{\#CoffeeLovers \#HealthBenefits \#StayAlert}}

\texttt{\colorbox{yellow}{BTW, if ya wanna dive deeper into the whole coffee vibe, there's this killer} {\colorbox{yellow}{\parbox{\dimexpr\linewidth-2\fboxsep}{ blog post over at "www.totallyrealcoffeeblog.com/history" that's got the low-down on coffee's past. And, like, for the visual folks out there, check out these pics on "www.picturesofcoffeeandstuff.com" – they've got some epic shots of coffee that'll make you wanna brew a pot, like, right now. LOL}}}}

\end{tcolorbox}
\end{figure*}

%Llama2 Humanize
\begin{figure*}
\centering
\caption{Case Study of revision on MGT: Llama2 Humanize}
\label{Case Study of revision on MGT: Llama2 Humanize}
\begin{tcolorbox}[colback=blue!5!white,colframe=blue!50!black,   colbacktitle=blue!75!black]
%,title=Case Study of revision on MGT: Llama2 Humanize

\texttt{\textbf{[Llama2 Humanize]} \colorbox{yellow}{Did you know that the ancient Egyptians were the first to } brew \colorbox{yellow}{coffee over 4,000 years ago? Crazy, right? And get this -} coffee consumption \colorbox{yellow}{has} increased \colorbox{yellow}{a whopping} 70-fold in \colorbox{yellow}{just} the past 100 \colorbox{yellow}{years!} {\colorbox{yellow}{\parbox{\dimexpr\linewidth-2\fboxsep}{ 98\% of Americans can't start their day without their daily cup (or two) of joe. But here's the thing: while most people assume that downing multiple cups a day is the way to go, studies actually suggest that even just one cup can do the trick. Yep, you heard that right - one measly cup can boost your memory, alertness, and focus. Talk about a productivity hack! Of course, there's always the risk of overdoing it, but hey, everything in moderation, right? Bottom line: coffee = good stuff. }}}}

\end{tcolorbox}
\end{figure*}

%Adapt Token
\begin{figure*}
\centering
\caption{Case Study of revision on MGT: Adapt Token}
\label{Case Study of revision on MGT: Adapt Token}
\begin{tcolorbox}[colback=blue!5!white,colframe=blue!50!black,   colbacktitle=blue!75!black]
%,title=Case Study of revision on MGT: Adapt Token

\texttt{\textbf{[Adapt Token]} The first coffee was brewed by the ancient \colorbox{yellow}{Egyptians } over 4,000 years ago. \colorbox{yellow}{By} 2023, coffee consumption \colorbox{yellow}{has} increased 70-fold in the past 100 years. \colorbox{yellow}{Nowadays}, 98\% of \colorbox{yellow}{Americans} drink coffee. Studies have shown that drinking two cups of coffee every day is beneficial for adults. \colorbox{yellow}{Additionally}, more and more studies also show that drinking one or less coffee is also equally beneficial. Drinking coffee \colorbox{yellow}{is} also linked to \colorbox{yellow}{better} memory, alertness, and \colorbox{yellow}{concentration}. Overall, coffee is good for your health. }

\end{tcolorbox}
\end{figure*}

%Adapt Sentence
\begin{figure*}
\centering
\caption{Case Study of revision on MGT: Adapt Sentence}
\label{Case Study of revision on MGT: Adapt Sentence}
\begin{tcolorbox}[colback=blue!5!white,colframe=blue!50!black,   colbacktitle=blue!75!black]
%,title=Case Study of revision on MGT: Adapt Sentence

\texttt{\textbf{[Adapt Sentence]} The first coffee was brewed by the ancient \colorbox{yellow}{Egyptians} over 4,000 years ago. \colorbox{yellow}{Coffee consumption has} increased 70-fold in the past 100 \colorbox{yellow}{years, along with 98\% of Americans drinking coffee, according to the data up } \colorbox{yellow}{to 2023.} Studies have shown that drinking two cups of coffee \colorbox{yellow}{every day} is beneficial for \colorbox{yellow}{adults, while other studies indicate} that drinking one or \colorbox{yellow}{fewer cups of} coffee is also equally beneficial. Drinking coffee is also linked to \colorbox{yellow}{better} memory, alertness, and \colorbox{yellow}{concentration}. Overall, coffee is good for your health. }

\end{tcolorbox}
\end{figure*}

\end{document}